\documentclass[iicol,sn-mathphys]{sn-jnl}
\usepackage{paralist}
\usepackage{comment}
\usepackage{cleveref}
\usepackage{amssymb}
\usepackage{graphicx}
\usepackage{subcaption}
\usepackage{psfrag}
\DeclareMathOperator{\sign}{sign}

\jyear{2022}%

\theoremstyle{thmstyleone}%

\theoremstyle{thmstyletwo}%
\newtheorem{remark}{Remark}%

\theoremstyle{thmstylethree}%
\newcommand{\pv}{p^{\textrm{m}}}
\newcommand{\ps}{p^{\textrm{s}}}
\newcommand{\ai}{\textrm{a}_i}
\newcommand{\Np}{N^{\textrm{p}}(\kappa)}
\newcommand{\xa}{x^{\textrm{a}}}
\newcommand{\ya}{y^{\textrm{a}}}
\newcommand{\xv}{x^{\textrm{v}}}
\newcommand{\yv}{y^{\textrm{v}}}

\begin{document}

\title[paper title]{Hierarchical Integration of Model Predictive and Fuzzy Logic Control 
for Combined Coverage and Target-Oriented Search-and-Rescue 
via Robots with Imperfect Sensors}

\author{\fnm{Christopher} \sur{de Koning}}
\author*{\fnm{Anahita} \sur{Jamshidnejad*}}
\email{A.Jamshidnejad@tudelft.nl}
\affil{\orgdiv{Control and Operations Department}, \orgname{Delft University of Technology}, \orgaddress{\city{Delft}, 
\country{the Netherlands}}}


\abstract{
Search-and-rescue (SaR) in unknown environments is a crucial task with life-threatening risks that 
requires precise, optimal, and fast decisions. Robots are promising candidates for autonomously 
performing various SaR tasks in unknown environments. While humans use their heuristics to effectively 
deal with uncertainties of SaR, optimisation of multiple objectives (e.g., mission time, area covered, 
number of victims detected) in the presence of physical and control constraints is a mathematical challenge 
that requires machine computations. Thus having both human-inspired and mathematical decision making capabilities is highly desired for SaR robots, while control approaches that exhibit both capabilities have been ignored significantly in the literature.  
Moreover, coordinating the individual decisions of robots with 
little computation cost in large-scale SaR missions is an open challenge. Finally, in real-life SaR missions 
due to defects (e.g., in sensors) or due to environmental factors (e.g., smoke) data perceived by SaR robots 
may be prone to uncertainties. We introduce a hierarchical multi-agent control architecture that exploits 
non-homogeneous and imperfect perception capabilities of SaR robots, as well as the computational efficiency 
and robustness to failure of decentralised control methods and global performance improvement of centralised 
control methods. The integrated structure of the proposed control framework allows to combine human-inspired 
and mathematical decision making methods, via respectively fuzzy logic and model predictive control, in a 
coordinated and computationally efficient way. The results of various computer-based simulations show that 
while the area coverage of the proposed approach is comparable to existing heuristic methods that are 
particularly developed for coverage-oriented SaR, the efficiency of the introduced approach in locating 
the trapped victims is significantly higher. Furthermore, with comparable computation times, the proposed 
control approach successfully avoids potential conflicts that exist in non-cooperative methods. These results 
confirm that the proposed multi-agent control system is capable of combining coverage-oriented and 
target-oriented SaR in a balanced and coordinated way. 
}




\maketitle

\section{Introduction}
\label{sec:intro}
Search-and-rescue (SaR) robots are expected to take over life-threatening tasks, 
especially within initial stages of searching an unknown environment, 
in order to reduce the risks for the SaR crew. 
SaR robots can potentially reduce the crucial time of finding the trapped victims
 and allow human resources to be available for other tasks, e.g., logistics and assisting the 
detected  victims \cite{Casper,Coburn}.  
Robots can move through areas that are inaccessible to humans, 
gather information (e.g., about the location of victims, explosive materials, and debris) 
and make maps of the environment. 
This way SaR robots contribute to improving the situational awareness for SaR crews, 
which is essential for mitigating the mission risks and for saving the lives 
of trapped victims \cite{Riley_human-robot,Shimanski2005,Chandarana2021}.%

SaR approaches can be categorised as target-oriented and coverage-oriented 
based on their control objectives. 
In target-oriented SaR, knowledge about the target distribution in the environment is 
initially available (see, e.g., \cite{Jamshidnejad2018,Beck_nonhomocollab,deAlcantaraAndrade_et_al_2019,sanjuan_fuzzyprioSARS,Yao_GBNN}). 
When the SaR environment is unknown, coverage-oriented approaches are mainly used  
\cite{Galceran_CPPsurvey}.   
Ant colony algorithms are bio-inspired area coverage methods that are computationally efficient 
and easy to implement \cite{Koenig_Ants,Wagner_ants}. 
Machine learning and neural network methods are also used for area coverage, 
where robots progressively learn effective area coverage behaviours \cite{Yang_ANNcoverage,yang_coopsearch}. 
The main drawback of such methods is their need for training before they can be implemented. 
Autonomous learning algorithms, including generalized model-free reinforcement learning methods, 
have thus been developed to address this challenge (see, e.g., \cite{Tutsoy2021,Tutsoy2017}). 
In these algorithms the system keeps on learning an optimal policy online. 
Although very promising, autonomous learning methods may face new challenges regarding 
computational burden for real-life implementations for SaR, due to the large size and varying dynamics 
of SaR environments that make the learning procedure more complicated. 
Moreover, in SaR missions there may be high risks associated with implementing a solely 
learning-based algorithm before the system achieves its optimal performance. 
More specifically, during the stages that the algorithm is learning an optimal policy, 
there are serious risks regarding losing the trapped victims or delaying their detection, 
which may result in their health state becoming critical.    
Moreover, most coverage-oriented approaches do not systematically  
incorporate victim or target detection in their search behaviour. 
Arnold et al.\ in \cite{Arnold_nonhomoSARS} present a cooperative, multi-agent 
SaR system with the objective of both victim detection and exploration   
in order to increase situational awareness of the environment. 
The SaR agents, however, are steered according to fixed behaviour sets. 
This limits the adaptability and thus efficiency of these robots in highly dynamic SaR environments. 
Existing SaR control methods are mainly focused on either coverage or target-oriented SaR. 
Moreover, MPC, which is an optimisation-based control method that systematically handles 
state and input constraints and can provide robustness to SaR uncertainties, has been ignored 
for the crucial task of area coverage in SaR \cite{deAlcantaraAndrade_et_al_2019}. 
Instead, MPC has mainly been used for reference tracking in target-oriented SaR in (partially) known environments (see, e.g., \cite{Jamshidnejad2018,farrokhsiar_et_al_2013,hoy_et_al_2012}).%

In order to speed up mapping the SaR area and to reduce the risk of mission failure, 
a fleet of SaR robots may be deployed (see, e.g., \cite{cooper_2020,deAlcantaraAndrade_et_al_2019,paez_et_al_2021}). 
In centralised SAR multi-agent control, robots are controlled via 
a centralised system that determines the mission plans for all these robots (see, e.g., 
\cite{Beck_nonhomocollab,deAlcantaraAndrade_et_al_2019,sanjuan_fuzzyprioSARS,Yao_GBNN}). 
In decentralised SaR multi-agent control, local (on-board) controllers 
are considered for robots 
(see \cite{Koenig_Ants,Yang_ANNcoverage,Arnold_nonhomoSARS,Choi_decentral,Liu_SARsurvey}.  
Best et al.\ \cite{Best2020} present a cooperative distributed information gathering approach for 
SaR robots where based on learning and heuristics robots visit stationary, pre-known goal regions. 
While a task assignment problem is solved in a communication-wise efficient way, 
there are no (dynamic) uncertainties involved in the environment of the robots. 
Otte et al.\ \cite{Otte2020} address a cooperative task-assignment problem using a decentralised auction approach. 
In particular, they investigate the effect of lossy communication among the agents on the performance of the multi-agent  
system with the aim of providing insight into the selection of an auction algorithm that, 
despite lossy communication, satisfies the desired performance criteria of a multi-agent system. 
A multi-agent search-planning approach is introduced in \cite{Kashino2020}  
for wilderness SaR with a team of aerial and ground robots. 
In their approach, the initial trajectory planning for the aerial robots is performed offline.  
After an aerial robot detects a (possibly moving) target, the robot tracks it until a ground robot  
intercepts this target.%

While decentralised control approaches are more robust to failure and are computationally more efficient than centralised approaches, providing reliable and stable communication among the robots and missing a global vision of the entire system are challenges of decentralised control approaches \cite{Choi_decentral}. 
Hierarchical architectures can combine the strengths of centralised and decentralised control approaches  
(see, e.g., \cite{Tol_Jamshidnejad_2021}). 
Particularly, for multi-robot SaR systems hierarchical control architectures can provide coordination in the behaviour of local controllers. However, a limited amount of research on hierarchical control for SaR robots is available. Examples include \cite{Khamis,Elston_hierarch,Chandler_hierarch}, which 
are all limited to target-oriented SaR.

Currently, SaR robots need (intensive) supervision and control from human operators. On the one hand, however, for safety, efficiency, and avoiding additional challenges regarding online human-robot interaction there is interest in making these robots autonomous  \cite{Liu_SARsurvey,Pellerin_surveySARS,Casper,Riley_human-robot}. 
On the other hand, humans use their heuristics effectively in order to deal with uncertainties of SaR missions. Therefore, providing SaR robots with human knowledge will improve their performance. Human knowledge, which is provided as information-based control for SaR robots in \cite{krzysiak_butail_2021}, was shown to improve the performance of SaR robots in finding the targets. 
Thus having both capabilities of human-inspired decision making and mathematical control 
is highly desired for SaR robots. However, control approaches that exhibit both capabilities 
have been ignored significantly in the literature.%

In this paper, we introduce a hierarchical control architecture for 
multi-agent control of SaR robots with non-homogeneous, imperfect sensors 
that combines mathematical and human-inspired control methods in a computationally efficient way. %
%
%
The main contributions of this paper include: 
\begin{enumerate}
    \item 
    Introducing a novel hierarchical control framework for multi-objective control and coordination of multi-robot systems and for exploiting their non-homogeneous sensor imperfections in unknown environments. 
    The resulting control system benefits both from robustness to failure and computational efficiency of decentralised 
    control methods and from globally effective performance of centralised control methods    
    \item 
    Integrating human-inspired and mathematical decision making by formulating local fuzzy logic controllers, which mimic decision making of human experts, and a supervisory model predictive control (MPC) system, which provides mathematical precision in the decisions of the system, systematically handles state and input constraints, improves the global performance of the multi-robot system based on its optimal and predictive decision making, and resolves conflicts of local heuristic controllers
    \item
    Implementing the proposed control approaches for combined coverage and target-oriented SaR via multi-robot systems with imperfect sensors for optimising the mission time, area coverage, and number of detected victims, and running extensive experiments via computer-based SaR simulations in order to evaluate various performance criteria (e.g., computational efficiency, percentage of area covered, overall certainty level of the map developed for the SaR environment, number of victims detected) of the proposed SaR control methods compared to the state-of-the-art methods  
\end{enumerate}
Additionally, since the local controllers steer the robots, the multi-robot control system will 
not fail due to a failure of the centralised controller. In that case, as our simulation results indicate, 
the remaining decentralised control system can still steer the SaR system safely, 
although with a degraded performance. Finally, MPC - which is an optimisation-based 
control method that systematically handles state and input constraints and that can 
provide robustness to SaR uncertainties - has been ignored in the literature for the 
crucial task of area coverage in SaR. Instead, MPC has mainly been used for reference tracking in target-oriented SaR in (partially) known environments. 
Our novel approach and formulation for multi-agent control systems enables MPC to 
provide all its strong points for, not only target-oriented, but also coverage-oriented SaR in unknown areas.%

The rest of the paper is structured as it follows. 
In \Cref{sec:problem} the problem formulation is detailed.  \Cref{sec:method} 
discusses the proposed hierarchical mission planning control approach for SaR robots. 
\Cref{sec:exp} describes the case study and experimental setup and  
presents, analyses, and discusses the results. 
\Cref{sec:concl} concludes the paper and provides suggestions for future research.%


\section{Problem Formulation}
\label{sec:problem}

In this section, we explain and formulate the details of the mission planning problem of 
SaR robots with non-homogeneous imperfect sensors. In particular, we discuss the modelling 
of the SaR environment and victims and the uncertainties involved,  
as well as the formulation of perception capabilities of SaR robots.%

\subsection{SaR Environment}
\label{subsec:env}

\begin{figure}
    \centering
    \includegraphics[width=0.48\textwidth]{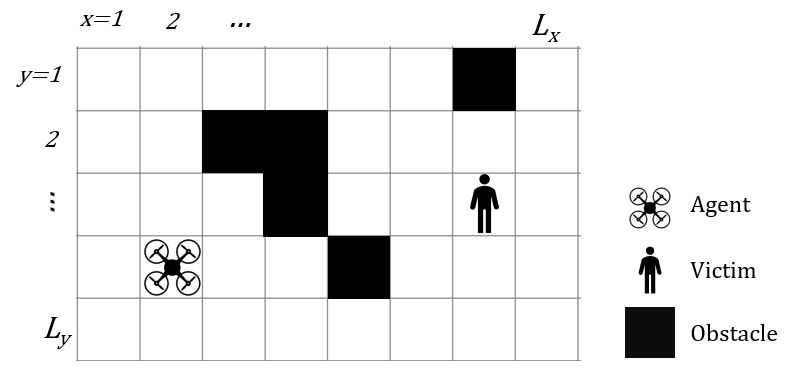}
    \caption{Schematic view of a SaR environment.}
    \label{fig:environment}
\end{figure}

The SaR environment $E$ is modelled by a bounded, discretised, 2-dimensional 
cellular area of $L_x \times L_y$ cells (see \Cref{fig:environment}.). 
Each cell in the SaR environment corresponds to the coordinates $(x,y)$ of its centre 
and may be empty or occupied by a static obstacle (i.e., wall, pillar, rubble), 
or by a victim and/or a SaR robot.  
A cell can embed a single victim at a time. Moreover, obstacles 
make a cell inaccessible for SaR robots and for victims. 
The following uncertainties for SaR robots have been considered: 
\begin{itemize}
    \item 
    External uncertainty regarding the SaR environment, i.e., the  
    total number of victims and obstacles and their positions are unknown
    \item
    External, random uncertainty about the pattern of movement of victims
    \item
    Internal (i.e., structural) and external (i.e., proximal) 
    uncertainties regarding the perceived data
\end{itemize}
An $L_x \times L_y$ matrix $\mathcal{W}(\kappa)$, called the \emph{occupancy map}, is used to 
record cells that are occupied by static obstacles after being detected by a SaR robot. 
Furthermore, whenever a victim is detected by the sensor of SaR robot $i$, 
the robot stores the location, 
perceived health state, and time of detection of the victim in 
a local matrix (specific to the robot) called the \emph{victim map} $\mathcal{V}_i(\kappa)$ 
of SaR robot $i$. This map is used by the controller of the robot 
to make the current control decision. 
However, for the sake of efficiency for the on-board computations and the memory storage, 
in their local victim maps the robots keep track only of those victims who 
have been selected as a target by the controller and have been visited by the robot 
(i.e., the robot has been in the same cell as the victim), 
as well as of those victims who are currently within the perception field of the robot's sensor. 
Thus SaR robots do not record any memory of victims who have previously been detected by the robot, 
but have not been selected as target for them.%

The scan certainty $c(x,y,\kappa)$ of cell $(x,y)$ for time step $\kappa$ 
is a value within $[0,1]$ that specifies the certainty level regarding the information available about cell $(x,y)$. 
Moreover, each cell $(x,y)$ corresponds to a proximal uncertainty  
at time step $\kappa$ that is a function of the Euclidean distance  
of the cell from all SaR robots that scan the cell. 
The scan certainty of a cell depends on whether or not the cell has been scanned by any SaR robots   
and if so, how accurate the perceived data is, i.e., the scan 
certainty depends on the proximal uncertainty.  
This relationship is explained in detail in Section~\ref{subsec:agents}.
Initially the scan certainty of all cells within the SaR environment is zero.  
The scan certainty for all cells is included in an $L_x \times L_y$ matrix $\mathcal{C}(\kappa)$, 
called the \emph{scan certainty map}, which will be updated in time.%


\subsection{SaR Robots}
\label{subsec:agents}

\begin{figure}
    \centering
    \includegraphics[width=0.45\textwidth]{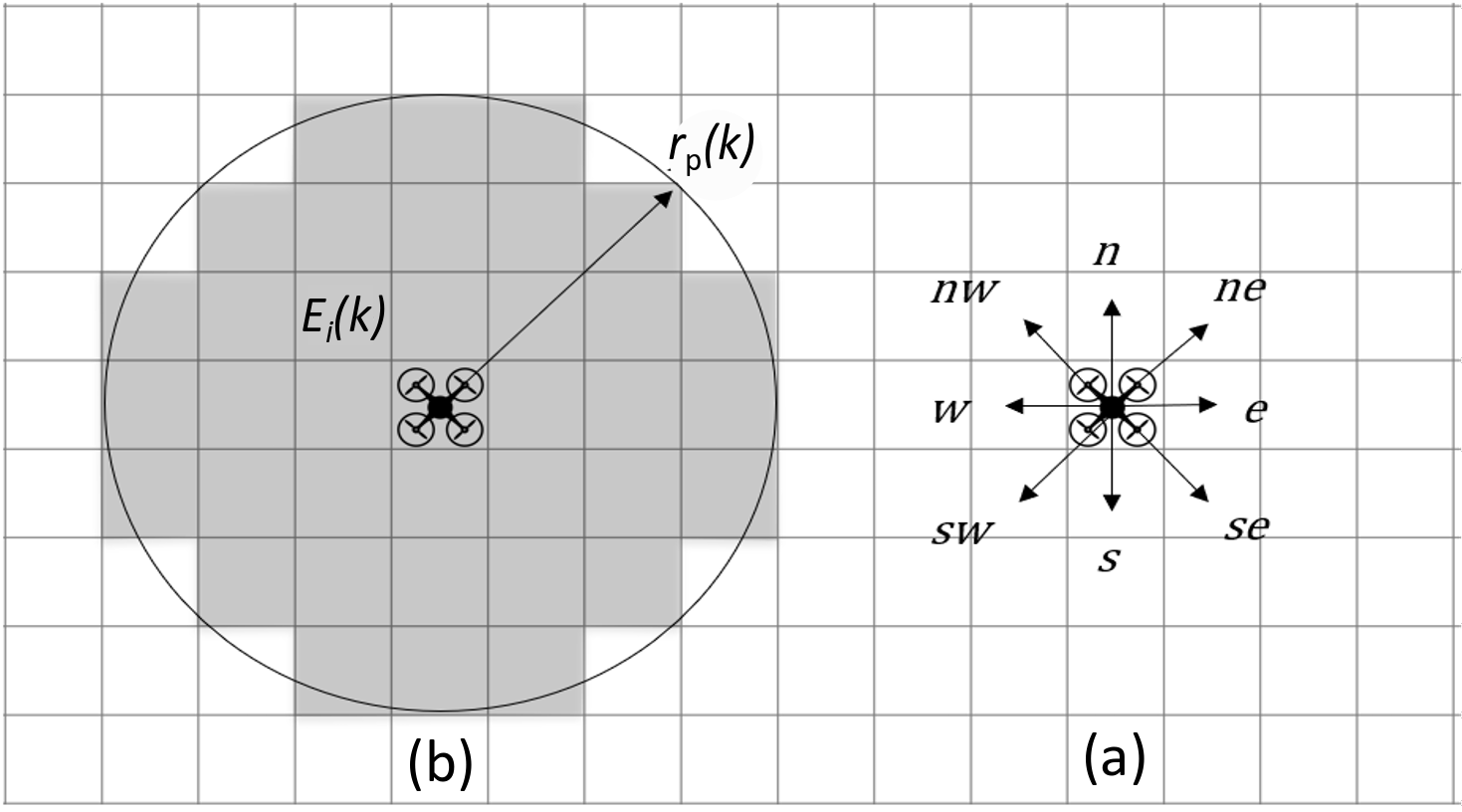}
    \psfrag{(a)}[][][]{(b)}
    \caption{(a) Movement possibilities of a SaR robot. (b) Perception field of a SaR robot.}
    \label{fig:percep}
\end{figure}

We consider a multi-robot SaR system composed of $N$ agents 
$\textrm{a}_i$ ($i=1,\ldots,N$) that, per simulation time step, 
may move to one of the $8$ neighbouring cells (see \Cref{fig:percep}(a)). 
These robots are equipped with optical cameras and sensors that localise the victims 
and that assess their health state 
(e.g., acoustic and heat sensors \cite{Casper,Ganesan} or sensors that detect 
WiFi-enabled devices \cite{Wang_wifiSARS}). 
The perception field $E_{i}(\kappa)$ of SaR robot $i$ for time step $\kappa$ 
includes all cells of the SaR environment that fall within a circle of radius 
$r_{\mathrm{p},i}$, centred at the position of the robot at  
time step $\kappa$, where $E_{i}(\kappa)\subseteq E$ (see Figure~\ref{fig:percep}(b)).%




The data perceived by SaR robots may in general be imperfect, 
i.e., scanning a cell does not necessarily yield full certainty about 
the information within the cell. 
Two sources of uncertainty regarding the perceived data are considered: 
(1) \emph{Structural imperfection}, which corresponds to a fixed perceptual uncertainty 
reduction rate $\eta_i \in (0,1]$ per SaR robot $\ai$. More specifically, 
every time SaR robot $\ai$ scans a cell, the uncertainty regarding the 
information of the cell is reduced by rate $\eta_i$. 
Thus when $\eta_i=1$ there is no structural imperfection 
(i.e., the perceived information corresponds to $100\%$ certainty). Moreover, 
we do not consider sensors that are completely out of function (i.e., $\eta_i=0$) 
due to structural imperfection. 
(2) \emph{Proximal uncertainty}, which implies that while all cells within the perception field 
$E_{i}$ of SaR robot $\ai$ are scanned, the degree of increase in the scan certainty 
of these cells decreases according to their distance from the sensor.%

The structural imperfection of sensors and the proximal uncertainty of the cells 
together will result in an uncertainty dynamic ratio $\sigma (x,y,\kappa)$    
corresponding to every cell $(x,y)$ per time step $\kappa$. We have: 
\begin{align}
\label{eq:dynamic_evolution_uncertainty}
z(x,y,\kappa + 1)  = \sigma(x,y,\kappa) z(x,y,\kappa)
\end{align}
with $z(x,y,\kappa)$  the scan uncertainty (i.e., $1 - c(x,y,\kappa)$) assigned 
to cell $(x,y)$ at time step $\kappa$, and:
\begin{align} 
\label{eq:rho}
    \sigma(x,y,\kappa) &= \prod_{i = 1} ^ N \sigma_i (x,y , \kappa)\\
\label{eq:rho_i}
    \sigma_i (x,y,\kappa)  & =  1 - (1 - \eta_i) e^{-r_i(x,y, \kappa)} \cdot \\
    & \frac{\displaystyle 1-\sign \Big(r_i(x,y, \kappa) -r_{\mathrm{p}, i} \Big)}{ \displaystyle 2} \nonumber
\end{align}
where $r_i(x,y, \kappa)$ is the Euclidean distance of SaR robot $\ai$ to cell $(x,y)$  
 at time step $\kappa$, $\sign(\cdot)$ represents the sign function, and  
$\sigma_i (x,y , \kappa) $ is the share of the uncertainty dynamic ratio of  
cell $(x , y)$ at time step $\kappa$ that is provided by the sensor of SaR robot $\ai$. 
The updated scan certainty for cell $(x , y)$ is given by: 
\begin{align}
    \label{eq:scan_certainty}
    c(x,y,\kappa + 1)  = 1 - z(x,y,\kappa + 1) 
\end{align} 
Based on \eqref{eq:rho_i}, the effect of the proximity on the uncertainty dynamic ratio  
of cells is modelled by an exponential function. 
More specifically, when $r_i(x,y, \kappa) = 0$, i.e., for the cell where SaR robot $\ai$ 
is currently located at, the uncertainty dynamic ratio corresponding to SaR robot $\ai$ is $\eta_i$ 
(i.e., the maximum possible improvement in the scan certainty of the cell that can be 
provided by the sensor of SaR robot $\ai$). 
This uncertainty dynamic ratio varies exponentially until for $r_i(x,y, \kappa) \geq r_{\mathrm{p}, i}$, 
it becomes unity (i.e., the scan certainty of the cell at the current time 
step does not improve as a result of a contribution of the sensor of SaR robot $\ai$).%

SaR robots may differ from each other in two properties regarding their sensors: 
(1) The sensors of SaR robots may have different perception radii $r_{\mathrm{p},i}$ 
for $i=1,\ldots,N$.   
(2) The accuracy of these sensors, and thus their perceptual uncertainty reduction rate $\eta_i$ 
may be different.%


\subsection{Victim Modelling}
\label{subsec:vic}

The number, location, and health state of the victims are initially unknown for the SaR robots. 
The victims follow a random pattern of movement, i.e.,  
victim $v$ with position $(\xv_v(\kappa),\yv_v(\kappa))$ 
at time step $\kappa$ may remain in its current cell with probability  
$\ps$ or may move to one of its (unblocked) neighbouring cells with a total probability $1-\ps$, 
which results in an equal probability of  
$\pv_v(\kappa) = 
(1 - \ps)/n^\textrm{free}(\xv_v(\kappa),\yv_v(\kappa),\kappa)$ 
to move to each of the (unblocked) neighbouring cells,  
where $n^\textrm{free}(x,y,\kappa)$ is the number of free neighbouring cells 
for cell $(x,y)$ at time step $\kappa$. Note that for the sake of simplicity we consider the probability 
$\ps$ to be constant in time, space, and for all victims. 
In case a more detailed model is to be used, this probability may vary with time, 
and per cell and victim.%

Moreover, each victim holds a certain health state, $h_{v}(\kappa)$, which varies within $[0,100]$ 
and implies how healthy or injured the victim is at time step $\kappa$. 
Whenever a victim is detected by a SaR robot, their initial health state is registered. 
Over time, the health state of each victim may decrease with the rate $\Delta h_v (\kappa)$ given by: 
\begin{align}
    \label{eq:viccond}
    {\Delta}h_v (\kappa) = 
    \bigg\{
    \begin{array}{cc}
     -\alpha                   & h^{\textrm{crit}}\leq h_v (\kappa)\leq 100  \\
     \beta h_v (\kappa) - \gamma & 0  \leq  h_v(\kappa) \leq h^{\textrm{crit}}
    \end{array}    
\end{align}
with $\alpha, \beta, \gamma > 0$, $h^{\textrm{crit}}$ the critical health state, 
and $\gamma \geq \beta h^{\textrm{crit}}$. 
Based on \eqref{eq:viccond}, a victim has a uniformly deteriorating health state  
whenever their health state is not less than $h^{\textrm{crit}}$ (i.e., health state is stable), 
while the rate of deterioration of the health state 
becomes linear as soon as the health state is below $h^{\textrm{crit}}$. 
The updated health state is given by:
\begin{align}
    h_v (\kappa + 1 ) = \max\left\{h_v (\kappa) + {\Delta}h_v (\kappa) , 0 \right\}
\end{align}
A SaR robot detects a victim whenever they are both in the same cell. 
Without considering the technical details regarding data analysis, sensor fusion, 
or soft sensing in this paper, we assume 
that the robot detects the victim (see, e.g., \cite{Dousai2022,Llasag2019}) 
and assesses the health state of the victim, e.g., using a combination of 
WiFi, optical, thermal, or acoustic sensors and using image processing algorithms 
or via direct feedback received from the victims when possible (see, e.g., \cite{Pinheiro2022}).%



\section{Hierarchical Control System}
\label{sec:method}

\begin{figure}
    \centering
    \includegraphics[width=0.35\textwidth]{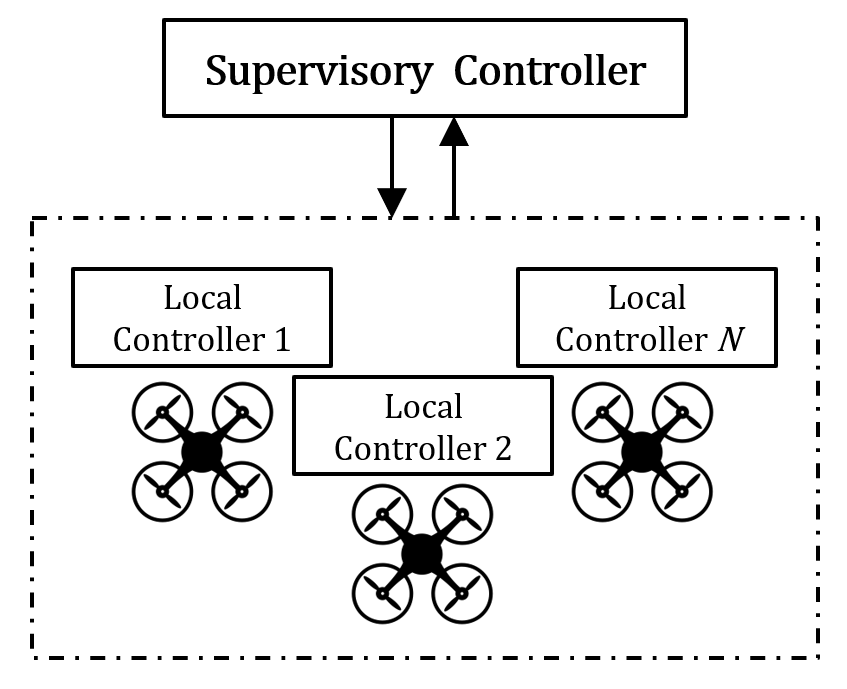}
    \caption{Hierarchical architecture of the proposed cooperative hierarchical mission planning controller.}
    \label{fig:hierarch}
\end{figure}

Next we explain the proposed hierarchical control system 
that steers the search behaviour of SaR robots. 
The control architecture includes two levels (see \Cref{fig:hierarch}):  
The lower level of control is composed of decentralised controllers that steer the local search behaviour of 
each SaR robot, while the higher control level includes a centralised supervisory controller that 
coordinates the behaviour of the decentralised controllers, such that search conflicts 
among SaR robots are resolved. 
SaR robots only communicate with the supervisory control level, 
without sharing any information among themselves. 
The proposed control architecture thus combines the strengths of centralised and decentralised control approaches.


\subsection{Local Fuzzy Logic Controllers}
\label{subsec:local}

At the local level, a SaR robot first processes the data that is captured via its sensors 
and then constructs a local priority map for its perception field. 
This map includes quantities corresponding to the importance of visiting the cells for the SaR robot.  
Next the local controller of the robot determines a path that yields the highest local gain 
according to a quantity called the path grade. 
Since the main objective of the SaR mission is to optimise the area coverage and 
time efficiency of detecting the victims, the following two main criteria are considered in grading a path: 
\begin{enumerate}
    \item \textbf{Time reduction:} 
    Each SaR robot should reach its targets in the least possible time 
    in order to contribute to minimising the overall mission time. 
    \item \textbf{Exploration increase:} 
    Each SaR robot should scan as many (unexplored) cells as possible along its path 
    in order to contribute to maximising the overall area coverage.
\end{enumerate}
These two criteria may possess a conflict, since for the first criterion the 
search behaviour should be target-oriented (in other words the robot should find the shortest 
possible path that leads it to the target as quickly as possible), while for the second criterion the 
search behaviour is coverage-oriented (in other words the robot should visit more cells before reaching its target). 
Therefore, the local controllers are developed such that a balanced trade-off between these criteria is provided.

\subsubsection{Search Priority Assignment}
\label{subsec:taskass}

The local controllers of SaR robots should first assign priorities to potential paths, 
specifying the urgency to scan every cell within the environment. 
For every cell in the perception field $E_i$ of SaR robot $\ai$, a priority score is determined 
using rule-based fuzzy logic control (FLC) methods. 
The main motivation for using FLC is its computational efficiency, 
which is essential for local controllers due to the limited computational power available on board for SaR robots, 
and the capability of FLC in mimicking human's logic for decision making, 
which allows to incorporate human expert knowledge in local controllers. 
Thus local controllers effectively mimic the reasoning of human experts 
without their direct supervision.%

Fuzzy rules with the following formulation are used by local controllers of SaR robots:
\begin{align}
    \label{eq:fisrules}
    \mathcal{R}_m: 
    & \text{\textbf{ If }} e^{\textrm{v}}(x,y,\kappa) \text{ is } A_{m,1}  \text{\textbf{ and }} h_v(\kappa) \text{ is } A_{m,2}  
    \\ 
     & \text{\textbf{ and }} c(x,y,\kappa) \text{ is } A_{m,3} \text{\textbf{ then }} \rho(x,y,\kappa) \text{ is } B_m \nonumber
\end{align}
with $e^{\textrm{v}}(x,y,\kappa)$ the probability of existing a victim in cell $(x,y)\in E_i$ 
at time step $\kappa$, 
$m = 1,\ldots, M$ with $M$ the number of rules, 
$A_{m,1}$, $A_{m,2}$, $A_{m,3}$, and $B_m$ 
fuzzy sets that adopt a linguistic term.%

The fuzzy inference system corresponding to the rules given by \eqref{eq:fisrules} 
receives $3$ inputs (i.e., the probability of existence of a victim in a cell, 
the health state of the potential victim, and 
the most recent scan certainty value of a cell) 
and assigns a search priority $\rho(x,y,\kappa)$ to cell $(x,y)$ for time step $\kappa$.  
Note that every SaR robot $\ai$ has access to its local knowledge stored in the local 
scan certainty map $\mathcal{C}_i(\kappa)$ and local victim map $\mathcal{V}_i(\kappa)$. 
The probability $e^{\textrm{v}}(x,y,\kappa)$ of existence of a victim in cell $(x,y)$ estimated by SaR robot $\ai$
depends on the robot's sensor, and adopts either a very small value when the sensor receives no signal 
that implies a victim exists in the cell (for a sensor with $\eta_i=1$ this small value may be $0$, 
while for a sensor with $\eta_i\in(0,1)$ a small positive value may be considered), 
or a percentage determined according to the structural imperfection and proximal  
uncertainty explained in Section~\ref{subsec:agents}. 
Based on \eqref{eq:fisrules}, cells within the perception field of the robot where it is likely to find a victim with a worse health state and has not yet (extensively) been scanned receive a higher priority.  
Inaccessible cells within the occupancy map $\mathcal{W}(\kappa)$ receive a null priority.


\subsubsection{Path Planning}
\label{subsec:pathplan}

After prioritising the cells, each local controller determines potential paths for the 
corresponding SaR robot. 
In order to optimise the time, shortest paths are favourable, while for optimising the area coverage, 
paths that visit more cells with higher priorities are preferred. 
Thus the local controller applies an A* search approach \cite{Hart_ASTAR} based on Yen's algorithm \cite{Yen_kshortest} 
to determine a certain number of shortest paths that end at every cell 
within the perception field of the robot. Afterwards these paths are graded based on 
their \textit{travel time} and \textit{degree of exploration}   
to specify how favourable they are for the SaR mission 
at the current time step. 
The travel time is computed based on the path length and the robot's speed. 
We suppose that a SaR robot moves one cell per time step, thus 
the travel time corresponds to the path length only. 
The degree of exploration of every potential path $P_i(\kappa) \subseteq E_{i}$ 
for SaR robot $\ai$ at time step $\kappa$ is computed via: 
\begin{align}
    \label{eq:return}
    \epsilon\left(P_i(\kappa)\right) = \sum_{k=\kappa }^{\kappa + \ell\left(P_i(\kappa)\right) - 1} 
    \lambda^k \rho(\xa_i(k),\ya_i(k), k)
\end{align}
where the path is defined by: 
\begin{align}
\label{eq:pathdef}
P_i&(\kappa) = \Big\{\left(\xa_i(\kappa),\ya_i(\kappa)\right),\ldots,\\ 
&\left(\xa_i\left(\kappa + \ell(P_i(\kappa))-1\right) , 
       \ya_i\left(\kappa + \ell(P_i(\kappa))-1\right)\right) 
       \Big\}
\nonumber
\end{align}
with $\ell(P_i(\kappa))$ the path length, $\lambda \in [0,1]$ the discount factor, 
and $\rho(\xa_i(k),\ya_i(k),k)$ the priority value of cell $(\xa_i(k),\ya_i(k))$, which the robot 
should visit at time step $k = \kappa,\ldots, \kappa + \ell(P_i(\kappa)) - 1$ 
when it follows path $P_i(\kappa)$. 
Note that since the priority values for the cells corresponding to time steps 
$k>\kappa$ are based on predicted estimates, considering a discount factor 
can reduce the potential influence of errors in the predictions. 
Finally, the grade of path $P_i(\kappa)$ is computed by (with $c_1,c_2>0$ constant values):
\begin{equation}
    \label{eq:grade}
    g\left(P_i(\kappa)\right) = -c_1 \ell\left(P_i(\kappa)\right) + c_2  
    \epsilon\left(P_i(\kappa)\right) 
\end{equation}
\begin{remark}
Since the paths that will be generated by the control system of the SaR robots 
are rectilinear, for practical implementations and to make the path easy to execute 
for real robots, we propose smoothening the paths before implementation (see, e.g.,  
\cite{Jamshidnejad_DeSchutter2018} for equations that can be used to smoothen such paths).
\end{remark}

\subsection{Supervisory MPC Controller}
\label{subsec:supervisor}

At the supervisory level, a centralised MPC-based controller is used that receives the local information 
corresponding to each SaR robot and merges them to build up global maps of the current perception fields of 
the robots. 
Note that while robots erase the non-target victims from their local victim maps 
(see \Cref{subsec:env} for details), the global victim map keeps track of all locally perceived information. 
This is practically possible because the global maps are recorded on a remote computer station 
that is not restricted by computational and memory limits.%

The supervisory controller is called  
whenever a search conflict is identified, i.e., the cardinality of the intersection of the perception fields of 
two SaR robots exceeds a certain threshold: 
$
\textrm{card}\left( E_{i} \cap E_j \right) 
> \tau_{\textrm{int}}
$. 
A model of the environment including the most updated cognitive maps is used 
as the prediction model of the supervisory controller, which determines  
globally optimal (within the controller's prediction time window) paths for SaR robots. 
This optimality is defined as a trade-off among various objectives including 
the mission time, the area coverage, and the chances of visiting more trapped victims with 
a more crucial health state. 
Despite providing globally optimal solutions, the MPC controller is computationally demanding 
due to the size of the centralised optimisation problem and the non-linearities involved in the problem. 
Therefore, we provide the supervisory controller with the paths that are determined by 
the local controllers as a warm start for the MPC optimisation problem  
to converge faster to an optimal solution. 
Taking into account the  objectives of the SaR mission, 
the objective function to be maximised by the supervisory controller at time step $\kappa$ is given by: 
\begin{align}
    \label{eq:obj}
    &J\left(\mathbb{P}(\kappa)\right) = \nonumber\\
    &w_1 \sum_{i=1}^{N} 
     g\left(P_i(\kappa)\right) + w_2 \sum_{(x,y)\in E} c \left(x,y, \Np \right)
\end{align}
with $\mathbb{P}(\kappa)$ (the optimisation variable) the set of paths for all the $N$ SaR robots 
and $w_1$ and $w_2$ constant weights. 
The objective function given by \eqref{eq:obj} is a weighted sum of two terms: 
(i) the overall grade of all paths (estimated by \eqref{eq:grade}) and 
(ii) the total predicted scan certainty of the SaR environment 
at the end of the current prediction horizon $\Np$, 
which is given by $\Np = \max_{i=1,\ldots,N}{\ell(P_i(\kappa))}$. 
Thus the second term steers the fleet of the SaR robots to spread out over the environment. 
In other words, the supervisory controller provides a balanced trade-off between locally preferred paths per robot and  
globally optimal paths from the point of area coverage. 
The supervisory controller does this using a global scan certainty map $\mathcal{C}(\kappa)$ 
of the environment and a global victim map $\mathcal{V}(\kappa)$, which are built by merging the local maps 
of SaR robots.

The supervisory control optimisation problem for time step $\kappa$ is given by 
(where the prediction window is $\{\kappa, \ldots, \Np - 1\}$): 
\begin{subequations}
\begin{align}
        & \max_{\mathbb{P}(\kappa)}J(\mathbb{P}(\kappa)) \nonumber\\
        &\textrm{such that:} \nonumber \\ 
         & \label{con:pathfeas} \mathbb{P}(\kappa)=\left\{P_1(\kappa), \ldots P_N(\kappa)\right\}  \\
     & \label{con:pathlegit} 
     \textrm{For all the paths, \eqref{eq:pathdef} holds, with } \\
     & \hspace{15ex}\left(\xa_i(\kappa),\ya_i(\kappa)\right)\in E\setminus\mathcal{W}(\kappa)  
     \nonumber\\
      \label{con:pathinter} 
     &\left(\xv_v(\kappa),\yv_v(\kappa)\right)\notin P_i(k)\cap P_j(k), \textrm{ where}\\ 
     & \hspace{7ex} i, j =1,\dots N, i \neq j, v = 1,\ldots,N^{\textrm{v}}(\kappa)\nonumber \\
     & \label{con:pathstart} \left(\xa_i(k),\ya_i(k)\right) = 
     \left(x^*_{i}(\kappa),y^*_{i}(\kappa)\right) \qquad  i=1,\dots N 
\end{align}
\end{subequations}
Constraints \eqref{con:pathfeas} and \eqref{con:pathlegit} define the optimisation variable 
and state that the paths should be feasible. 
Constraint \eqref{con:pathinter} restricts multiple SaR robots to  visit the same victim, where 
$N^{\textrm{v}}(\kappa)$ is the number of victims detected until simulation time step $\kappa$. 
This constraint improves the victim search efficiency and area coverage of the robots. 
To reduce the conservativeness of the problem and avoid infeasibility, 
constraint  \eqref{con:pathinter} may be defined as a chance (instead of a hard) constraint. 
Finally, constraint \eqref{con:pathstart} allows the starting point of the paths to be the 
most recent measured coordinates $(x^*_{i}(\kappa),y^*_{i}(\kappa))$ of the corresponding SaR robot.
\begin{remark}
Since the objective function of the supervisory MPC-based controller is defined in 
\eqref{eq:obj} as a weighted sum of the multiple control objectives, these objective terms 
will be normalised when implementing the optimisation problem. 
\end{remark}
%

\section{Case Study}
\label{sec:exp}

Next we discuss the results of computer-based simulations that are systematically designed 
to evaluate the performance of the proposed hierarchical control approach 
in comparison with state-of-the-art approaches for SaR. 
The simulations are implemented via MATLAB R2019b on a PC with Intel Core i7 
Processor with $2.20$ GHz frequency.   
Whenever an optimization problem should be solved to determine the paths of the SaR robots, 
the path planning problem is solved using pattern search as optimization method, 
since this algorithm showed to be faster than other alternative approaches. 
In order to make sure that the resulting paths meet the requirements of a discrete 
cellular environment for the numerical simulations (i.e., the way points defining 
the path have to be located at the centre of a cell) the continuous coordinates 
for the way points determined by pattern search are projected to the centre of the 
cells using the round function. For the parameters of the algorithms, 
we did a manual tuning with respect to the default settings.

\subsection{Simulation Setup}
\label{subsec:comparison}

We consider the following four search approaches that are common for SaR and compare their performance, 
in terms of victim detection, area coverage, and computational efficiency,  
with the proposed hierarchical control approach, which we call \textbf{cooperative controller} 
due to the supervisory MPC level. 
\textbf{Selfish controller:} A control system composed of the local controllers  
described in \Cref{subsec:local}, where the main difference with the 
cooperative controller is the lack of a supervisory controller. 
These controllers make decisions that fit their own circumstances only. 
\textbf{Pure MPC controller:} An optimisation-based search approach with the MPC 
structure of the supervisory controller described in \Cref{subsec:supervisor}, where the main difference with the cooperative 
controller is the lack of warm starting with trajectories that are proposed by the local controllers. 
Instead, as it is common in the implementation of MPC, the MPC controller 
receives the shifted solution of the previous time step as a warm start (see \cite{Diehl2005} for more details).  
Note that, in order to account for non-convexity of the problem, we have also run the simulations for pure MPC with multiple starting points within the given time budget. However, the results for pure MPC with warm start were still better. Thus only the results regarding pure MPC with warm start have been presented in the paper.%

\noindent
\textbf{ACS controller:} A heuristic ant-colony-based search approach based on \cite{Koenig_Ants}, where  
the global scan certainty map $\mathcal{C}(\kappa)$ is used for pheromone map for the ant colony system.%

\noindent 
\textbf{Exhaustive controller:} A random search strategy for SaR robots, 
where such search strategies are commonly used as reference bases for the other search methods.%

\begin{table}
    \centering
    \caption{Modelling and control parameters}
    \label{tab:env}
    \begin{tabular}{lrl}
    \toprule
    \textbf{Parameter} & \textbf{Value} & \\ \midrule
    $L_x$          & 40 &             \\
    $L_y$          & 25 &             \\
    Number of victims    & 25 &             \\
    $\ps$              & 0.6        \\
    $\alpha$           & 0.25   &          \\
    $\beta$           & 1/60   &          \\
    $\gamma$           & 1   &          \\
    $h^{\textrm{crit}}$ & 30 &   \\
    $\lambda$           & 0.6   &          \\
    $c_1$           & 2.0   &          \\
    $c_2$            & 5.0    &         \\
    $\tau_{\textrm{int}}$      & 30     &         \\
    $w_1$              & 1.0    &         \\
    $w_2$              & 0.05    &        \\ 
    \botrule
    \end{tabular}
    \end{table}
\begin{table}
    \centering
    \caption{Rule base of the fuzzy inference system (with the inputs given by  $e^{\textrm{v}}$, i.e., the probability of existence of a victim, 
    $h_v$, i.e., the health state of the potential victim, and $c$, i.e., the scan certainty of the cell to be visited, and the output is $\rho$, i.e., the search priority of the cell)}
    \resizebox{0.48\textwidth}{!}{\begin{tabular}{l|lll|l}
       \hline
        $\mathcal{R}_m$ & $e^{\textrm{v}}$ & $h_v$ & $c$ & $\rho$ \\\hline
    	1 &    Low & Stable & Known & Very Low \\       
        2 &    Low & Medium & Known & Very Low \\      
        3 &    Low & Stable & Partial & Very Low \\      
        4 &    Low & Medium & Partial & Low \\     
        5 &    Low & Critical & Known & Low \\        
        6 &    Medium & Medium & Partial & Low \\       
        7 &    Medium & Critical & Known & Low \\       
        8 &    Low & Stable & Unknown & Low \\          
        9 &    Medium & Stable & Known & Medium \\      
        10 &   Medium & Medium & Known & Medium \\    
        11 &   Medium & Stable & Partial & Medium \\    
        12 &   High & Stable & Partial & Medium \\      
        13 &   High & Medium & Known & Medium \\        
        14 &   High & Critical & Known & Medium \\      
        15 &   Low & Critical & Partial & Medium \\       
        16 &   Low & Medium & Unknown & Medium \\       
        17 &   High & Stable & Known & High \\          
        18 &   Medium & Stable & Unknown & High \\      
        19 &   Medium & Medium & Unknown & High \\      
        20 &   High & Stable & Unknown & High \\        
        21 &   Low & Critical & Unknown & High \\       
        22 &   Medium & Critical & Partial & Very High \\
        23 &   High & Medium & Partial & Very High \\    
        24 &   Medium & Critical & Unknown & Very High \\
        25 &   High & Medium & Unknown & Very High \\    
        26 &   High & Critical & Partial & Very High \\  
        27 &   High & Critical & Unknown & Very High \\\hline
    \end{tabular}}
    \label{tab:fisrules}
\end{table}

\begin{table}
    \centering
    \caption{Parameters of SaR robots}
    \label{tab:agent}
    \resizebox{0.48\textwidth}{!}{\begin{tabular}{lrrrllrrr}
    \cline{1-4} \cline{6-9}
    \textbf{General case} & \multicolumn{1}{l}{$r_{\mathrm{p}, i}$} & \multicolumn{1}{l}{$\eta_i$} & \multicolumn{1}{l}{$(\xa_{i,0},\ya_{i,0})$} &  & 
    \textbf{Case 3}                                               & \multicolumn{1}{l}{$r_{\mathrm{p}, i}$} & \multicolumn{1}{l}{$\eta_i$} & \multicolumn{1}{l}{$(\xa_{i,0},\ya_{i,0})$} \\ \cline{1-4} \cline{6-9} 
    $i = 1$  & 6                         & 0.1                        & (1,16)                               &  
    & $i = 1$                                                & 7                         & 0.1                        & (9,10)                               \\
    $i = 2$  & 4                         & 0.3                        & (13,25)                              &  & 
    $i = 2$                                                & 3                         & 0.3                        & (5,8)                                \\ \cline{1-4} \cline{6-9} 
                    & \multicolumn{1}{l}{}      & \multicolumn{1}{l}{}       & \multicolumn{1}{l}{}                 &  &                                                               & \multicolumn{1}{l}{}      & \multicolumn{1}{l}{}       & \multicolumn{1}{l}{}                 \\ \cline{1-4} \cline{6-9} 
    \textbf{Case 1} & \multicolumn{1}{l}{$r_{\mathrm{p}, i}$} & \multicolumn{1}{l}{$\eta_i$} & \multicolumn{1}{l}{$(\xa_{i,0},\ya_{i,0})$} &  
    & \textbf{Case 4}                                               
    & \multicolumn{1}{l}{$r_{\mathrm{p}, i}$} & \multicolumn{1}{l}{$\eta_i$} & \multicolumn{1}{l}{$(\xa_{i,0},\ya_{i,0})$} \\ \cline{1-4} \cline{6-9} 
    $i = 1$  & 6                         & 0.1                        & (10,8)                               &  & 
    $i = 1$                                                & 4                         & 0.1                        & (8,7)                                \\
    $i = 2$  & 4                         & 0.3                        & (10,6)                               &  & 
    $i = 2$                                                & 4                         & 0.3                        & (8,9)                                \\ \cline{1-4} \cline{6-9} 
                    & \multicolumn{1}{l}{}      & \multicolumn{1}{l}{}       & \multicolumn{1}{l}{}                 &  &                                                               & \multicolumn{1}{l}{}      & \multicolumn{1}{l}{}       
                    & \multicolumn{1}{l}{}                 \\ \cline{1-4} \cline{6-9} 
    \textbf{Case 2} & \multicolumn{1}{l}{$r_{\mathrm{p}, i}$} & \multicolumn{1}{l}{$\eta_i$} & \multicolumn{1}{l}{$(\xa_{i,0},\ya_{i,0})$} &  & \textbf{\begin{tabular}[c]{@{}l@{}} Case 5\end{tabular}} & \multicolumn{1}{l}{$r_{\mathrm{p}, i}$} & \multicolumn{1}{l}{$\eta_i$} & \multicolumn{1}{l}{$(\xa_{i,0},\ya_{i,0})$} \\ \cline{1-4} \cline{6-9} 
    $i = 1$  & 6                         & 0.1                        & (6,10)                               &  & 
    $i = 1$                                                & 7                         & 0.1                        & (9,17)                               \\
    $i = 2$  & 4                         & 0.3                        & (6,8)                                &  & 
    $i = 2$                                                & 3                         & 0.3                        & (5,15)                               \\ \cline{1-4} \cline{6-9} 
    \end{tabular}}
\end{table}

\begin{figure*}
    \centering
    \includegraphics[width=0.83\textwidth]{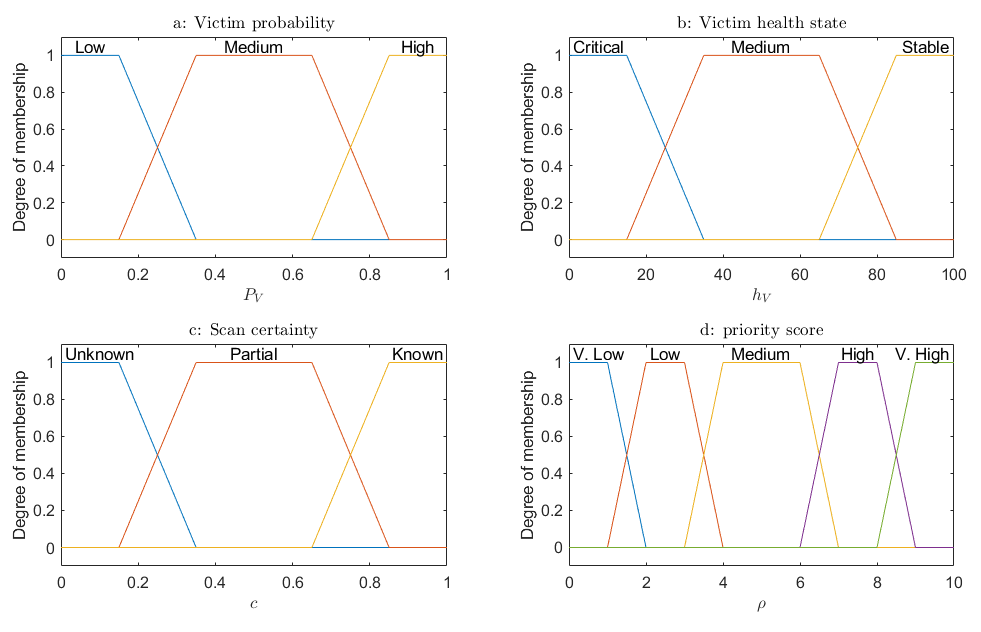}
    \caption{Trapezoidal fuzzy membership functions defined for the inputs and outputs of the Mamdani rule bases. 
    }
    \label{fig:mfs}
\end{figure*}

\begin{remark}
Note that, ideally, a centralised MPC controller can provide the desired performance for a system  
by providing a globally optimal solution. 
This requires to provide enough computational resources and time for the centralised MPC controller.
However, a main challenge that needs to be addressed for SaR problems is to provide a balanced trade-off 
between performance and computation time, such that the developed control system can meet the real-time 
requirements of a SaR robotic team. 
Therefore, we are interested in assessing how well different control approaches can steer the behaviour 
and performance of the SaR system when they are constrained by the computation time. 
Thus for both the pure MPC controller and the supervisory MPC controller 
we have considered a limited time budget, which may in some cases imply 
a degradation of the performance to meet the given computation time. 
\end{remark}

A set of $20$ simulation scenarios, each lasting $300$ simulation time steps, 
with a seeded random placement of victims and obstacles  
in an environment of a fixed size is considered. 
The parameters required for these simulations to estimate the movement of the victims and to  
compute  \eqref{eq:viccond} and \eqref{eq:return} are given in \Cref{tab:env}. 
The coefficients/weights in \eqref{eq:grade} and \eqref{eq:obj} 
are also given in \Cref{tab:env}, where the corresponding values are tuned manually via 
extensive experiments. 
The terms that describe the sets $A_{m,1}$, $A_{m,2}$, $A_{m,3}$, and $B_{m}$ in 
\eqref{eq:fisrules} should for real-life scenarios be deduced from real 
expert knowledge. 
For the numerical simulations designed in our case studies, we have defined 
the corresponding rule base based on intuition. 
More specifically, the sets $A_{m,1}$, $A_{m,2}$, and $A_{m,3}$ 
are verbally described by, respectively, ``Low, Medium, High'', ``Critical, Medium, Stable'', 
and ``Unknown, Partially Known, Known''. 
This selection allows us to build up a Mamdani rule base composed of $27$ rules. 
For the output set $B_{m}$ we select one of the following terms,  
``Very Low'', ``Low'', ``Medium'', ``High'',  ``Very High'',  
based on intuition and suited for the given realisations of the input fuzzy sets.  
The resulting Mamdani rule base is represented in \Cref{tab:fisrules}. 
The corresponding membership functions used in \eqref{eq:fisrules} are shown in \Cref{fig:mfs},
where trapezoidal functions have been selected, since they have proven to result 
in good quality control systems in various real-life applications (see \cite{FMF} for details).%

For the case study, we consider $2$ SaR robots with different sensory perception radii and 
perceptual uncertainty reduction rates. 
The robots start at fixed coordinates without any initial information about the SaR environment, 
thus all maps are initialised to zero/null. 
The parameters used for the SaR robots are shown in \Cref{tab:agent}. 
These parameters have been selected such that the influence of non-homogeneity 
and imperfection of the sensors can properly be incorporated into the numerical simulations.%

In order to evaluate various search approaches in terms of area coverage, victim search efficiency, and computational efficiency, 
the following performance metrics are considered. 
The area coverage is assessed via two performance metrics: (1) Total scan certainty of the environment as a function of simulation time steps, i.e.: 
\begin{equation}
    S(\kappa) = \sum_{\forall (x,y)\in E} c(x,y,\kappa)
\end{equation}
(2) Rise time for the total scan certainty (of a particular percentage). 
These performance metrics quantify the absolute area coverage, as well as 
the speed, thus efficiency, of each search approach. 
The victim search efficiency is evaluated via three metrics: (1) Number of (live and deceased) victims 
detected per simulation time step. (2) Simulation time step for which each victim is detected. 
(3) Health state of each victim at the time of detection. 
Finally, the average time for making control decisions per simulation time step is used to report the computational effort of each control approach.

\subsection{Results \& Discussions}
\label{subsec:results}

\begin{figure}
    \centering
%
    \includegraphics[width=0.5\textwidth]{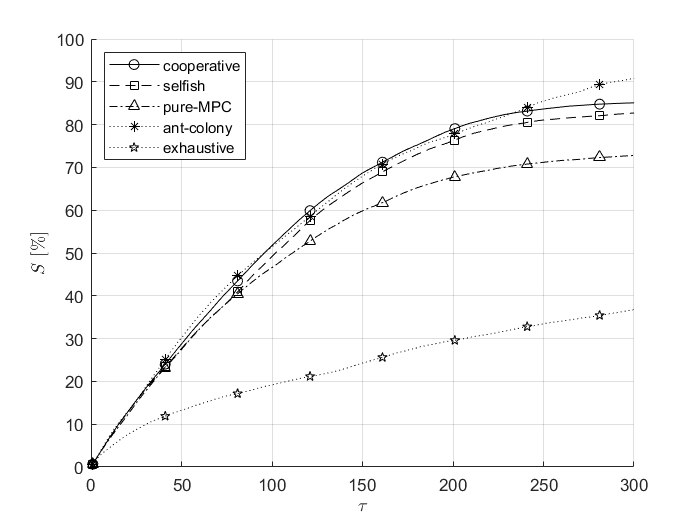}
    \caption{Total scan certainty corresponding to different controllers 
    as a percentage of the maximum scan certainty that can be obtained for the SaR environment.}
    \label{fig:scanmap}
\end{figure}

\begin{figure*} 
    \includegraphics[width=0.9\textwidth]{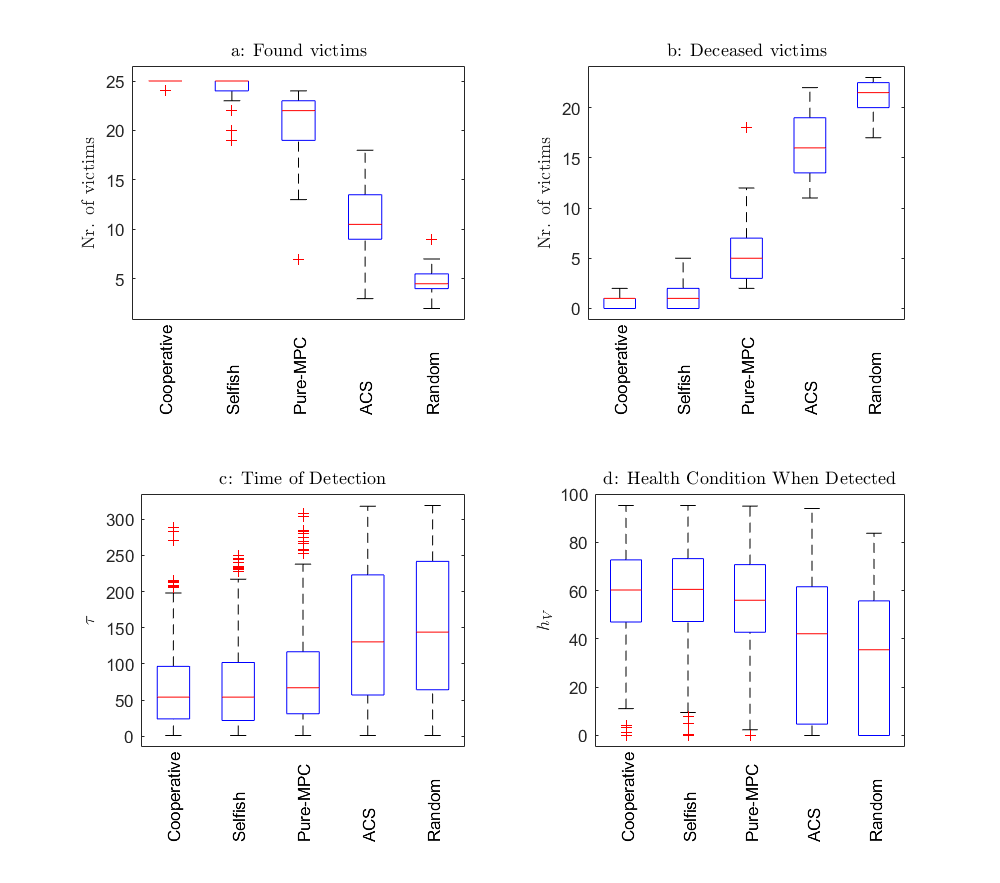}
    \caption{Victim detection efficiency for each control approach evaluated by (a) the number of victims found, 
    (b) the number of victims deceased, (c) time of detection, and (d) health state of victims when detected.}
    \label{fig:vicdetec}
\end{figure*}

\begin{table}
    \centering
    \caption{Number of simulation time steps needed to reach certain degrees of scan certainty.}
    \label{tab:risetime}
    \resizebox{0.48\textwidth}{!}{\begin{tabular}{lrrrrr} 
    \toprule
       Degree of scan certainty                  & \multicolumn{1}{l}{$50\%$} & \multicolumn{1}{l}{\textbf{$70\%$}} & 
       \multicolumn{1}{l}{\textbf{$80\%$}} & \multicolumn{1}{l} 
       {\textbf{$85\%$}} & \multicolumn{1}{l}{\textbf{$90\%$}} \\\midrule
    \textbf{Cooperative controller} & 97                                          & 157                                                  & 208                                                  & \multicolumn{1}{r}{291}                              & -                                                    \\
    \textbf{Selfish controller}     & 104                                         & 167                                                  & 238                                                  & -                                                    & -                                                    \\
    \textbf{ACS controller}         & 97                                          & 159                                                  & 218                                                  & \multicolumn{1}{r}{249}                              & \multicolumn{1}{r}{292}                              \\
    \textbf{Pure MPC controller}    & 112                                         & 229                                                  & \multicolumn{1}{c}{-}                                & -                                                    & -                                                    \\
    \textbf{Exhaustive controller}      & \multicolumn{1}{c}{-}                       & \multicolumn{1}{c}{-}                                & \multicolumn{1}{c}{-}                                & -                                                    & -                                                   \\\botrule
    \end{tabular}}
\bigskip
    \centering
    \caption{Average computation time for decision making per simulation time step for different controllers.}
    \resizebox{0.48\textwidth}{!}{\begin{tabular}{ccccc}
        \toprule
        \textbf{Cooperative} & \textbf{Selfish} & \textbf{ACS} & \textbf{Pure MPC} &\textbf{Exhaustive} \\
        \textbf{controller} & \textbf{controller} & \textbf{controller} & \textbf{controller} &\textbf{controller} \\\midrule
         4.5 [s] & 3.2 [s] & ~0 [s] & 8.6 [s] & ~0 [s] \\\botrule
    \end{tabular}}
    \label{tab:comp}
\end{table}

\begin{figure}
    \centering
    \includegraphics[width=0.45\textwidth]{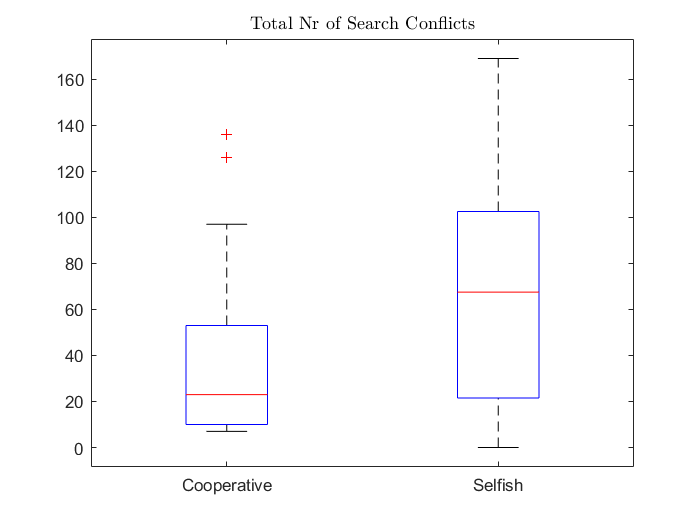}
    \caption{Number of simulation time steps when conflict is detected for cooperative and selfish controllers.}
    \label{fig:confl}
\end{figure}

\Cref{fig:scanmap}, \Cref{tab:risetime}, \Cref{fig:vicdetec}, and \Cref{tab:comp} represent the 
results of the simulations including, respectively, the total scan certainty, 
the number of simulation time steps required per control approach in order to 
reach a total scan certainty of $50\%, 70\%, 80\%, 85\%$, and $90\%$,  
the victim detection efficiency, and the average decision making time. 
Moreover, \Cref{fig:confl} shows the number of simulation time steps when search conflicts 
is registered for the cooperative and selfish controllers. 
For the cooperative controller, the number of registered conflicts is the number of times the supervisory 
MPC controller has been activated.%

\Cref{fig:scanmap} shows that the selfish and cooperative controllers achieve a 
comparable total scan certainty of above $80\%$ by the end of the simulations (to be more accurate, 
with a slightly higher percentage (around $85\%$) for the cooperative controller). 
However, \Cref{tab:risetime} shows that the cooperative controller is significantly faster 
 in reaching particular levels of scan certainty 
in later stages of the simulation 
(e.g., needs $14.4\%$ less time to reach an $80\%$ total scan certainty). 
The ACS controller (see \Cref{fig:scanmap}) reaches an overall scan certainty that is $6.71\%$ larger than 
that of the cooperative controller, with a comparable rise time in earlier stages of simulation. 
This is because the objective function of the ACS controller solely considers the scan certainty map 
to determine the most favourable next step for each SaR robot per time step. 
Thus the ACS controller has a single, non-conflicting objective as opposed to the selfish and cooperative controllers. 
Based on \Cref{fig:scanmap} and \Cref{tab:risetime}, the pure MPC controller performs much better than the 
exhaustive controllers, but worse than the cooperative, selfish, and ACS controllers. 
This is mainly because the pure  MPC approach is prone to falling within a sequence of 
local optima since it relies on the solution of the previous time steps as warm start. 
This not only highlights the importance of using more systematic (i.e., in line with the 
global objectives of the SaR mission) warm starts for the MPC controller, 
but also stresses that the exploratory nature of the selfish and cooperative controllers 
plays an important role in avoiding such issues.  
Finally, the exhaustive controller shows the worst performance 
regarding the area coverage, due to its lack of systematic search objective.%

Figures~\ref{fig:vicdetec}(a) and \ref{fig:vicdetec}(b) show that 
the cooperative and selfish controllers 
achieve a similar number of $25$ victims found and using these 
controllers correspond to the least number of victims deceased. 
However, considering all the  simulation runs, 
 the cooperative controller has lower variances, i.e., $0.134$ and $0.345$, 
for respectively the number of victims found and deceased than those 
(i.e., $3.10$ and $2.22$) of the selfish controller. 
This indicates a more consistently satisfactory performance 
for victim search using the cooperative controller. 
Both the selfish and cooperative controllers outperform the pure MPC, ACS, and exhaustive controllers  
in terms of victim detection, with the ACS and the exhaustive controller showing the worst performance 
(see Figures~\ref{fig:vicdetec}(a) and \ref{fig:vicdetec}(b)). 
This is because systematic victim detection is not an objective for these controllers. 
More specifically, with the ACS and exhaustive controllers SaR robots may detect the victims randomly. 
The fact that the ACS controller detects more victims than the exhaustive one is an indirect 
influence of its higher area coverage (see Figure~\ref{fig:scanmap}). 
While the pure MPC method outperforms both the ACS and  exhaustive controllers, 
compared to the cooperative and selfish controllers less victims are detected and more 
victims are deceased. This is due to the lower area coverage by the pure MPC controller, 
which has a negative impact on the victim detection efficiency. 
Based on Figures~\ref{fig:vicdetec}(c) and \ref{fig:vicdetec}(d) the cooperative and selfish controllers  
perform equally well considering the detection time and health state of victims. 
While the MPC controller detects less victims than the cooperative and selfish 
controllers, the detection time and health state of the victims found 
is at a comparable level (i.e., it is slightly worse) 
as those of the cooperative and selfish controllers.%

Finally, the computation time per decision (see \Cref{tab:comp})  
for the cooperative controller (i.e., $4.5$~s) is 
almost half of the computation time when only MPC is used (i.e., pure MPC controller) to steer the system. 
Moreover, compared to the average decision making time of the selfish controller (i.e., $3.5$~s), 
and considering the significantly better performance of the 
cooperative controller, this controller is the best choice among all the given controllers. 
Based on \Cref{fig:confl}, compared to the selfish controller search conflicts 
happen less when the cooperative controller is used. 
Thus each time the supervisory controller is triggered, by improving the global 
performance of the SaR system, it reduces the number of future search conflicts.%

Based on the results and discussions given above, 
the cooperative controller significantly outperforms 
the other methods. The next best controller is the selfish controller, that is 
the decentralised control system that remains when the supervisory MPC controller 
is excluded. 
These results further confirm the robustness of the proposed control 
architecture to failure of the supervisory MPC controller, 
i.e., while the performance degrades after the supervisory MPC controller is excluded 
from the control architecture, the performance of the SaR robotic team is still 
better than the other control methods used in the case study.%

\subsection{Structured Simulation Scenarios}
\label{subsec:specials}

\begin{table*}
\centering
\caption{Victim detection results for the combined scenario (case 5). }
\label{tab:combi_vic}
\resizebox{.75\textwidth}{!}{\begin{tabular}{ccccccccc}
\toprule
                   &  &\textbf{Selfish controller}                        &                       &  &                    
                   & & \textbf{Cooperative controller}                   &                       \\
                   
                   \cmidrule{1-4}\cmidrule{6-9}
                  victim & health state & number of visits  & detection time step &  
                  & victim & health state & number of visits  & detection time step \\
                   \cmidrule{1-4}\cmidrule{6-9}
$\boldsymbol{\mathrm{v}_1}$ & 6.94                        & 0 & -                     &  & $\boldsymbol{\mathrm{v}_1}$ & 6.94                        & 0 & -                     \\
$\boldsymbol{\mathrm{v}_2}$ & 24.9                        & 2 & 10                    &  & $\boldsymbol{\mathrm{v}_2}$ & 17.2                        & 1 & 22                    \\
$\boldsymbol{V_3}$ & 17.97                       & 2 & 3                     &  & $\boldsymbol{V_3}$ & 14.34                       & 2 & 8                     \\
$\boldsymbol{V_4}$ & 0                           & 0 & -                     &  & $\boldsymbol{V_4}$ & 6.03                        & 1 & 11                    \\
$\boldsymbol{V_5}$ & 6.94                        & 0 & -                     &  & $\boldsymbol{V_5}$ & 21.87                       & 1 & 15                    \\
$\boldsymbol{V_6}$ & 6.94                        & 0 & -                     &  & $\boldsymbol{V_6}$ & 11.15                       & 1 & 30                    
\\\botrule
\end{tabular}}
\end{table*}

\begin{figure*}
    \begin{minipage}{0.48\textwidth}
        \centering
        \includegraphics[width=0.85\linewidth]{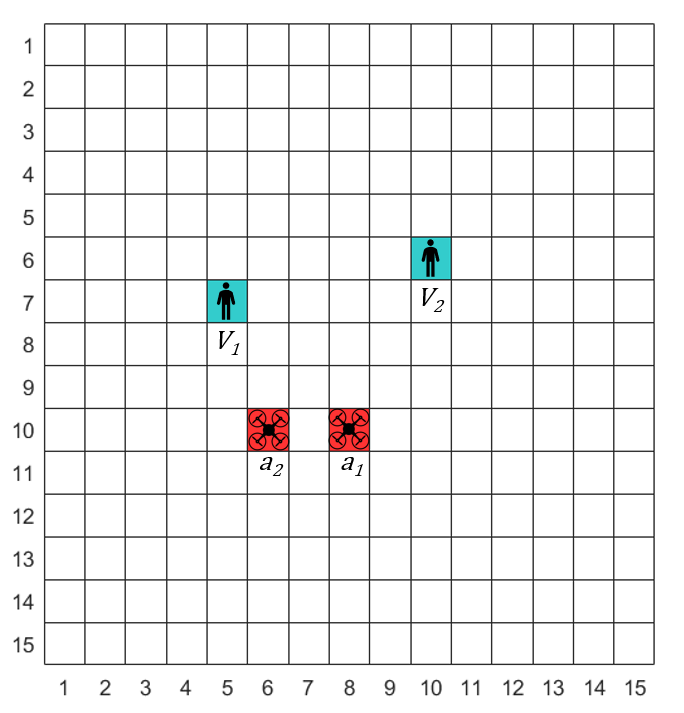}
        \caption{Case 1: Conflict in victim detection.}
        \label{fig:case1}
    \end{minipage} \quad
    \begin{minipage}{0.5\textwidth}
        \centering
        \includegraphics[width=0.85\linewidth]{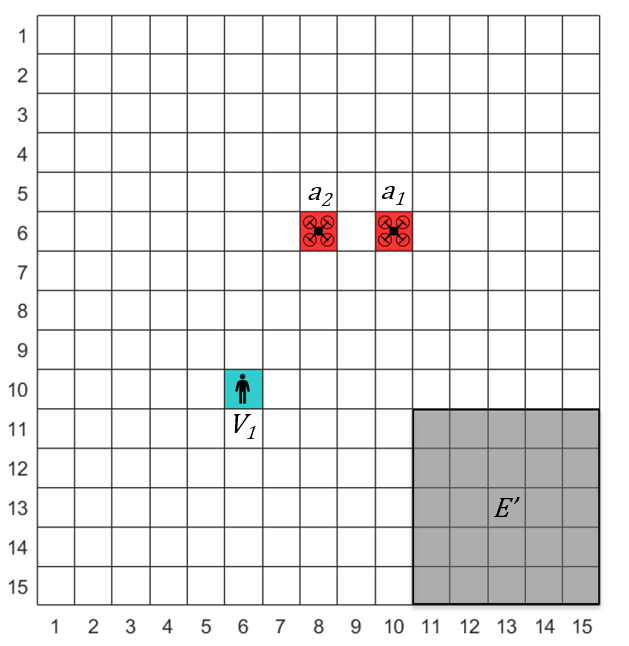}
        \caption{Case 2: Conflict in victim detection and area coverage.}
        \label{fig:case2}
    \end{minipage}
    \bigskip
    \begin{minipage}{0.48\textwidth}
        \centering
        \includegraphics[width=0.82\linewidth]{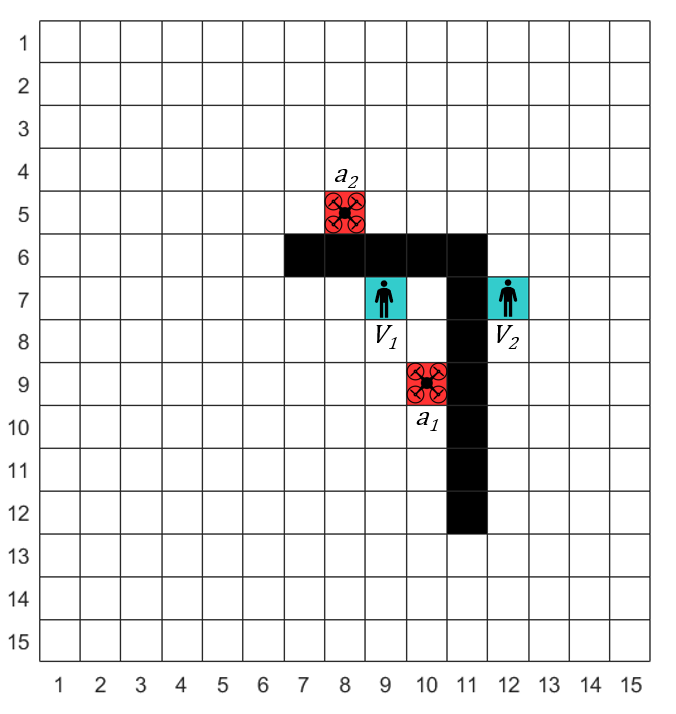}
        \caption{Case 3: Exploitation of perception fields.}
        \label{fig:case3}
    \end{minipage} \quad
    \begin{minipage}{0.5\textwidth}
        \centering
        \includegraphics[width=0.82\linewidth]{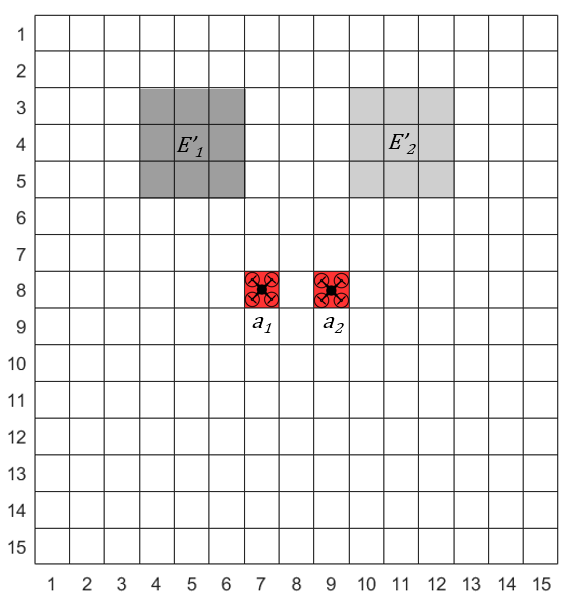}
        \caption{Case 4: Exploitation of sensor accuracies.}
        \label{fig:case4}
    \end{minipage}

    \centering
    \includegraphics[width=0.72\textwidth]{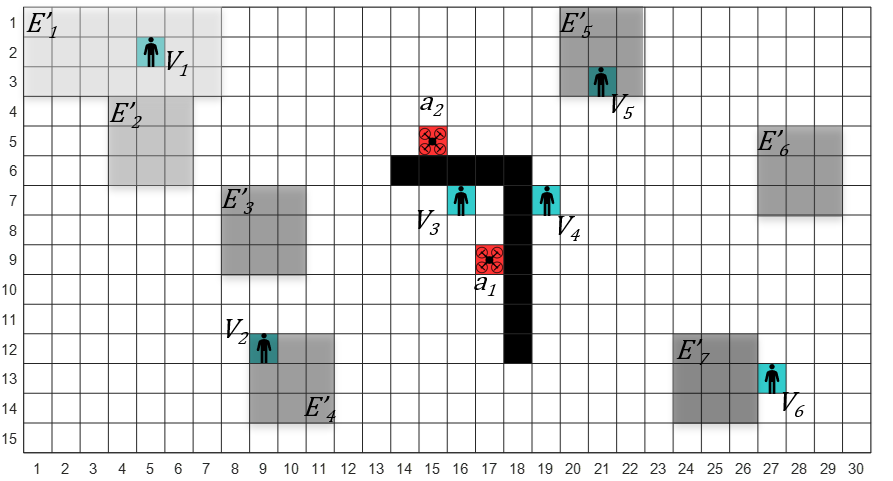}
    \caption{Case 5: Combined scenario.}
    \label{fig:combi_layout}
\end{figure*}

\begin{figure*}
    \centering
    \includegraphics[height=0.3\textheight]{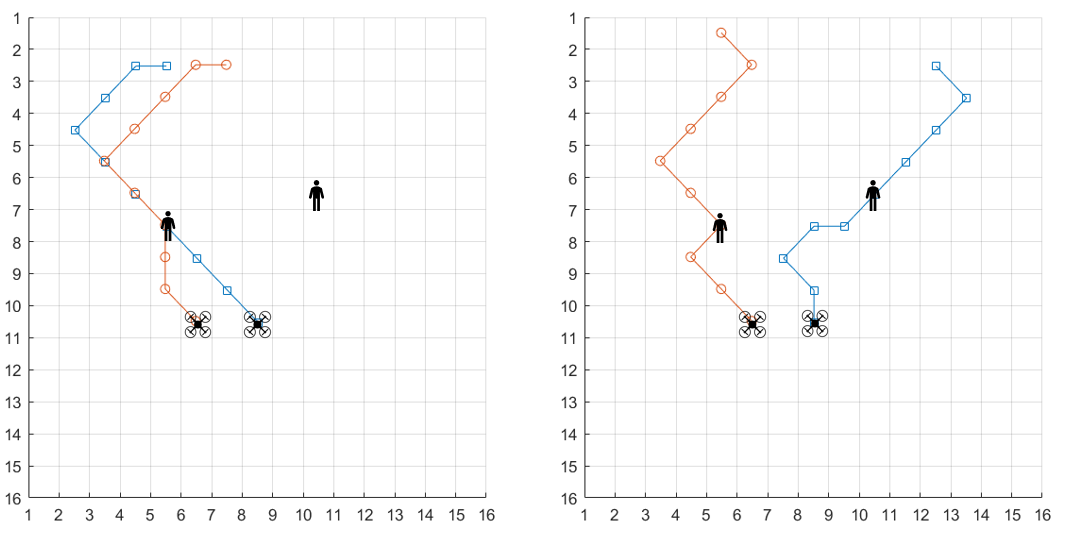}
    \caption{Case 1 - Path taken by SaR robots using the selfish controller (left) and the cooperative controller (right).}
    \label{fig:1track}
    \bigskip
    \centering
    \includegraphics[height=0.3\textheight]{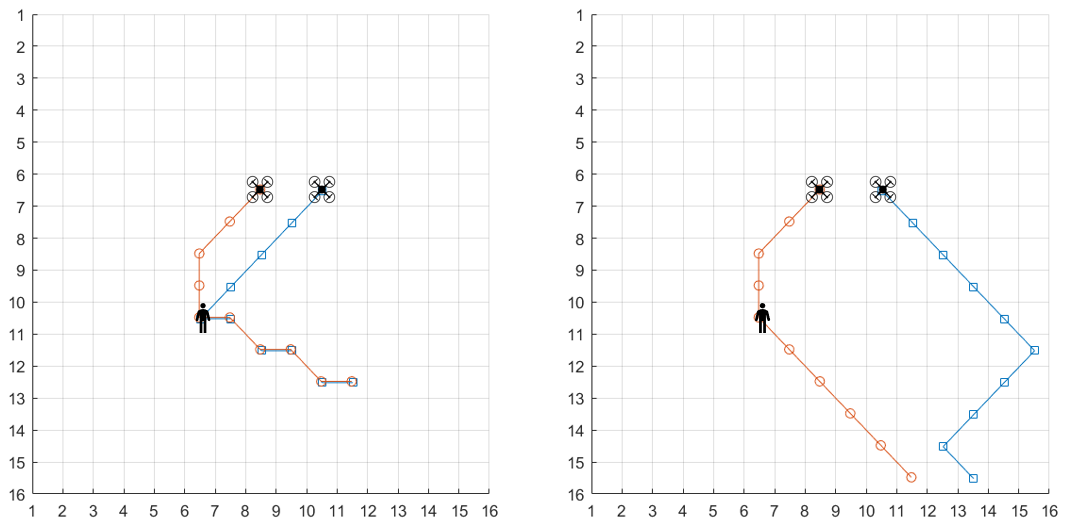}
    \caption{Case 2 - Path taken by SaR robots using the selfish controller (left) and the cooperative controller (right). }
    \label{fig:2track}
    \bigskip
    \centering
    \includegraphics[height=0.3\textheight]{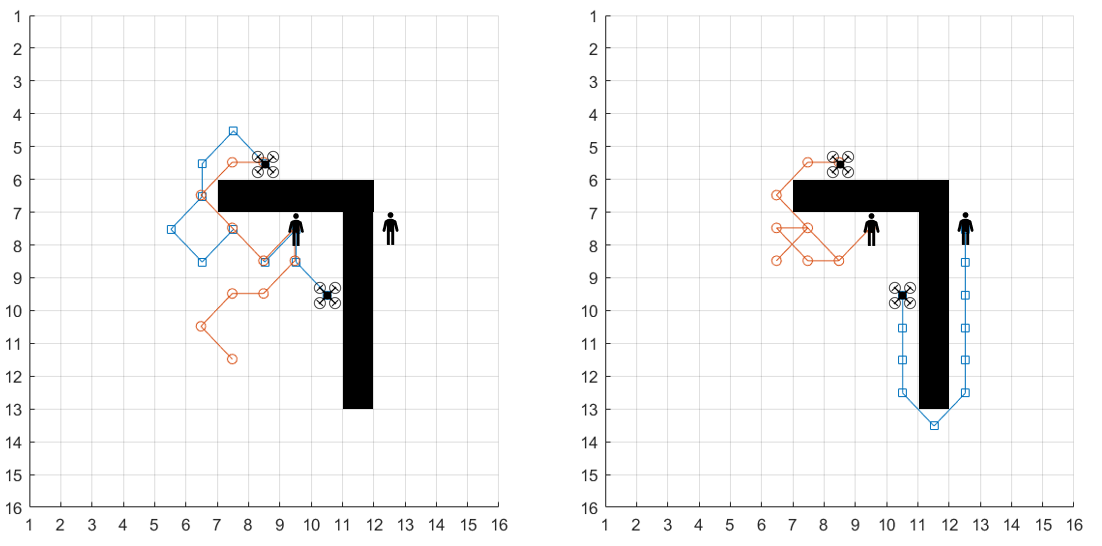}
    \caption{Case 3 - Path taken by SaR robots using the selfish controller (left) and the cooperative controller (right).}
    \label{fig:3track}
\end{figure*}

\begin{figure*}
    \centering
    \includegraphics[height=0.3\textheight]{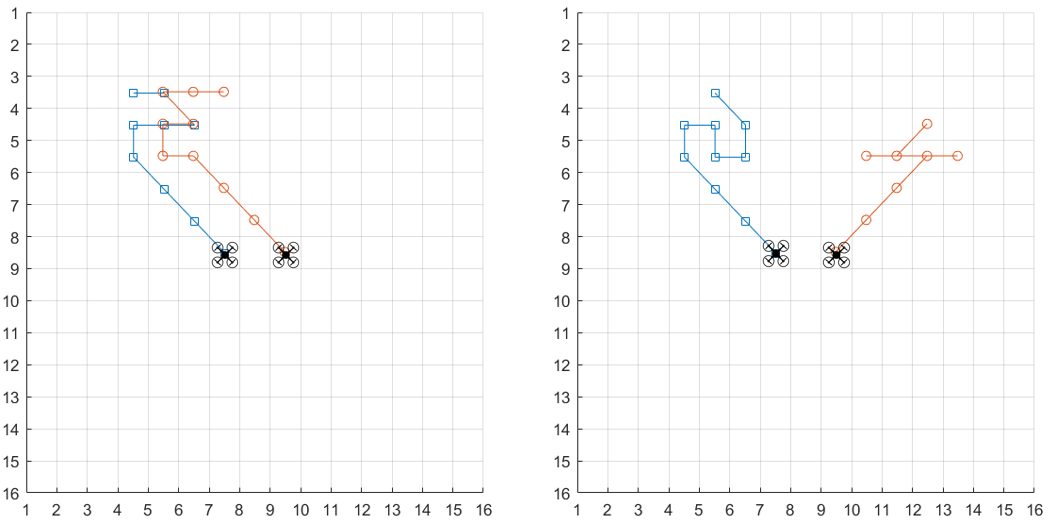}
    \caption{Case 4 - Path taken by SaR robots using the selfish controller (left) and the cooperative controller (right).}
    \label{fig:4track}
    \bigskip
    \centering
    \includegraphics[height=0.6\textheight]{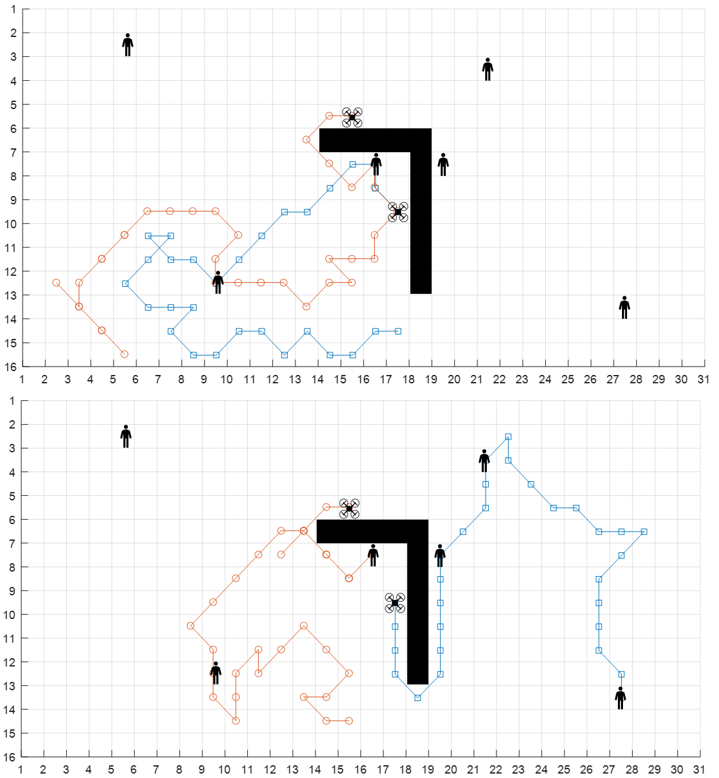}
    \caption{Case 5 - Path taken by SaR robots using the selfish controller (top) and the cooperative controller (bottom).}
    \label{fig:combitrack}
\end{figure*}

\begin{figure*}
    \begin{minipage}{0.4\textwidth}
        \centering
        \includegraphics[width=0.95\linewidth]{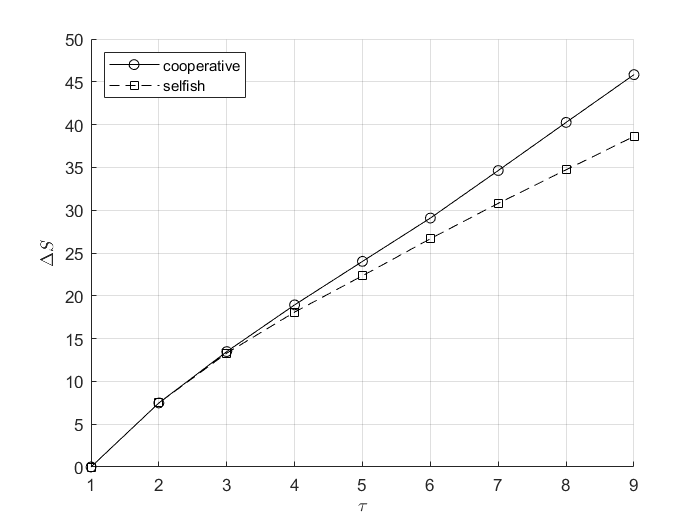}
        \caption{Case 1 - Evolution of the scan certainty in time.}
        \label{fig:1scan}
    \end{minipage} \quad
    \begin{minipage}{0.4\textwidth}
        \centering
        \includegraphics[width=0.95\linewidth]{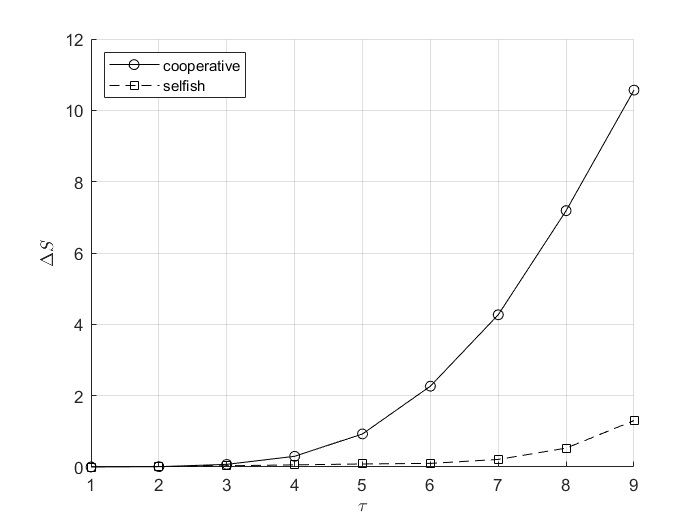}
        \caption{Case 2 - Evolution of the scan certainty in time.}
        \label{fig:2scan}
    \end{minipage}
    \bigskip
    \begin{minipage}{0.4\textwidth}
        \centering
        \includegraphics[width=0.95\linewidth]{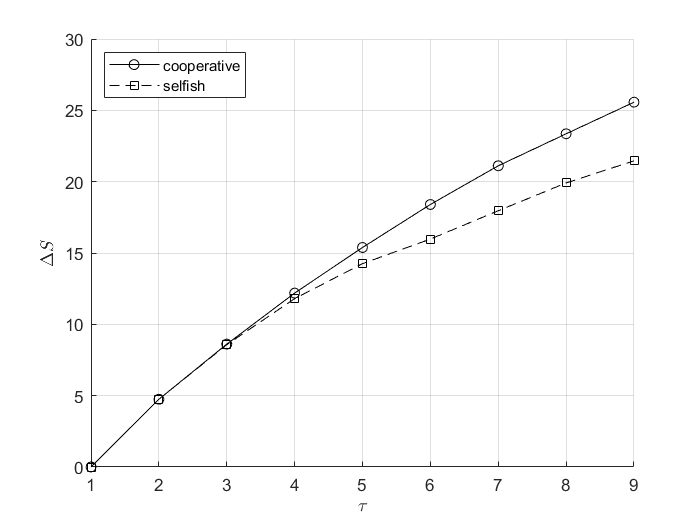}
        \caption{Case 3 - Evolution of the scan certainty in time.}
        \label{fig:3scan}
    \end{minipage} \quad
    \begin{minipage}{0.4\textwidth}
        \centering
        \includegraphics[width=0.95\linewidth]{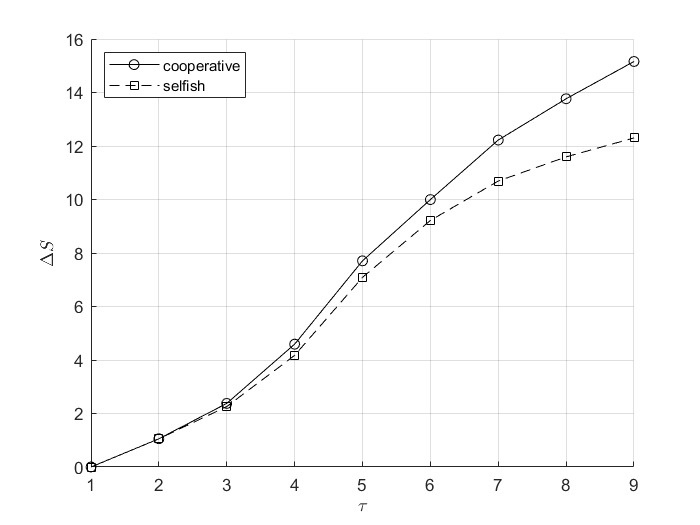}
        \caption{Case 4 - Evolution of the scan certainty in time.}
        \label{fig:4scan}
    \end{minipage}
    \bigskip
    \centering
    \includegraphics[width=0.4\linewidth]{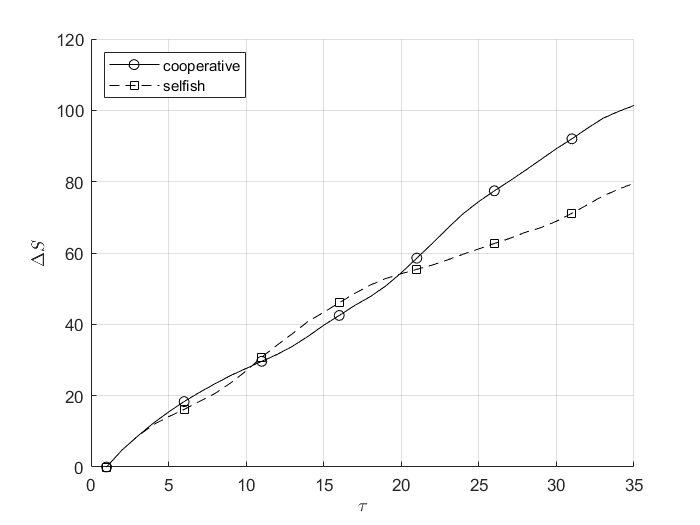}
    \caption{Case 5 - Evolution of the scan certainty in time. }
    \captionsetup{justification=justified}
    \label{fig:combiscan}
\end{figure*}

In order to further assess the performance of the  
cooperative controller in a more structured way and 
to assess the problem solving behaviour of the controller when 
several types of conflicts among the local controllers exist, 
five cases of conflicts in smaller scales than the previous simulation scenarios  
are considered (see Figures~\ref{fig:case1}-\ref{fig:combi_layout}).%

\noindent
\textbf{Case 1. Conflict in victim detection:}  
Consider $2$ SaR robots and $2$ victims in a partially known environment of size $15\times 15$  
with $c(x,y,0)=0.2$ for all $(x,y) \in E$ (see \Cref{fig:case1}). 
The health states of the victims are $10$ and $50$. 
Figures~\ref{fig:1track} and \ref{fig:1scan} show, respectively,   
the paths taken by the robots for $9$ time steps using  
the cooperative and the selfish control methods, and  
the change in the total scan certainty in time.%

Based on \Cref{fig:1track} with the selfish controller, both robots prioritise victim 
$\mathrm{v}_1$ over victim $\mathrm{v}_2$, since these robots are steered by local controllers that 
follow \eqref{eq:fisrules}-\eqref{eq:grade}, which prioritise visiting a cell that includes a victim with the worst health state and 
that is closer to the SaR robot 
(where the importance of each factor depends on the values for parameters $c_1$ and $c_2$). 
Since the cell that embeds victim $\mathrm{v}_1$ meets both conditions,  
victim $\mathrm{v}_1$ is the target of both SaR robots and is detected by them at time step $3$. 
Afterwards, the robots continue exploring the environment without moving to victim $\mathrm{v}_2$, 
who remains outside of their perception fields.%

With the cooperative controller, however, SaR robot $\textrm{a}_2$ detects victim $\mathrm{v}_1$ at  
time step $3$ and  SaR robot $\textrm{a}_1$ detects victim $\mathrm{v}_2$ at time step $5$ 
(see Figures~\ref{fig:case1} and \ref{fig:1track}). 
This shows that the supervisory MPC controller has successfully coordinated 
the actions of the SaR robots in favour of detecting more victims within a given time span. 
More specifically, the second term in the objective function of the MPC controller 
(see \eqref{eq:obj}) prevents the two SaR robots to cover the same sub-area 
of the environment. Since the local loss (considered by the first term in \eqref{eq:obj}) 
for redirecting SaR robot $\textrm{a}_1$ 
towards victim $\textrm{v}_2$ is less than that of redirecting the other robot, 
the supervisory MPC controller changes the path that has been 
proposed by the local controller of SaR robot $\textrm{a}_1$ in order to achieve a higher 
global gain considering the scan certainty of the environment. 
The overall scan certainty for the cooperative controller  
is $17.2\%$ larger than that of the selfish controller (see \Cref{fig:1scan}).%

\noindent
\textbf{Case 2. Conflict in victim detection and area coverage:} 
Consider $2$ SaR robots and $1$ victim in an environment of size $15\times 15$ 
that is known except for sub-area $E' \subset E$ that is completely unknown, 
i.e., $c(x,y,0)=0$ for all $(x,y) \in E'$ and $c(x,y,0)=1$ for all $(x,y) \in E \setminus E'$ 
(see  \Cref{fig:case2}). 
Figures~\ref{fig:2track} and \ref{fig:2scan} show, respectively, 
the paths taken by the SaR robots for $9$ time steps using the cooperative and selfish 
controllers and the change in the total scan certainty in time.%

\Cref{fig:2track} shows that for the selfish controller, both robots  
move towards victim $\mathrm{v}_1$ and visit the victim at time step $4$. 
Afterwards, both robots follow identical paths to approach the unknown sub-area. 
These individual behaviours are steered by the local controllers that, 
according to \eqref{eq:fisrules}-\eqref{eq:grade}, enforce each robot to prioritise cells 
that may embed a victim, are closer to the robot, and gain a higher percentage 
for the overall scan certainty of the environment. 
Since the cell that embeds the victim, in addition is closer to the robots than the cells 
within sub-area $E'$, this cell for the local controllers has a higher priority. 
After visiting the victim, based on \eqref{eq:fisrules}, visiting the closest cell 
within sub-area $E'$ becomes the priority of the local controllers.%

With the cooperative controller SaR robot $\textrm{a}_1$ that is farther 
from the victim is redirected via the supervisory MPC controller to move 
directly towards sub-area $E'$, while SaR robot $\textrm{a}_2$ moves towards victim 
$\mathrm{v}_1$ and visits the victim at time step $4$. 
With both the selfish and the cooperative controllers, victim $\mathrm{v}_1$ 
is detected equally fast, while based on \Cref{fig:2scan}, with the cooperative controller  
the overall scan certainty is almost $4$ times larger than the value for the selfish controller. 
This is an influence of including the second term of \eqref{eq:obj} in the objective function 
of the supervisory MPC controller.%

\noindent
\textbf{Case 3. Exploitation of perception fields:} 
Consider $2$ SaR robots and $2$ victims in a partially known environment 
of size $15\times 15$ with $c(x,y,0)=0.5$ for all $(x,y) \in E$ and a set of obstacles 
shown in black in \Cref{fig:case3}. 
SaR robot $\textrm{a}_1$ has a sufficiently large perception field such that it detects both victims, 
whereas SaR robot $\textrm{a}_2$ detects victim $\mathrm{v}_1$ only. 
The health states of victims  $\mathrm{v}_1$ and $\mathrm{v}_2$ are, respectively,  $20$ and $15$. 
Note that it is assumed that although the cells that include obstacles (shown in black) 
block the movement of the SaR robot, but they do not obstruct the view of the robot. 
Figures~\ref{fig:3track} and \ref{fig:3scan} show, respectively, 
the paths taken by the robots for $10$ time steps using the cooperative and 
selfish controllers and the change in the total scan certainty in time.%

With the selfish controller,  both SaR robots visit victim $\mathrm{v}_1$, 
which occurs at time steps $2$ and $5$ for robots 
$\textrm{a}_1$ and $\textrm{a}_2$, respectively (see \Cref{fig:3track}). 
Although the health state of victim $\mathrm{v}_1$ is less critical compared to 
that of  victim $\mathrm{v}_2$, victim $\mathrm{v}_1$ is prioritised over victim 
$\mathrm{v}_2$ by the local controller of SaR robot $\textrm{a}_2$, because this victim 
can be reached faster and thus the corresponding path receives a larger grade using \eqref{eq:grade}. 
Afterwards, the SaR robots continue exploring the environment without detecting victim $\mathrm{v}_2$ 
due to their limited perception fields.%

With the cooperative controller, SaR robots $\textrm{a}_1$ and $\textrm{a}_2$ 
find victims $\mathrm{v}_2$ and $\mathrm{v}_1$ at time steps $10$ and $5$, respectively. 
Although SaR robot $\textrm{a}_1$ is able to reach victim $\mathrm{v}_1$ in a shorter time, 
the global decision of the supervisory controller allows victim $\mathrm{v}_1$ to be detected later 
in order to make sure that victim $\mathrm{v}_2$ is detected in time 
(this is taken care of via the first term in \eqref{eq:obj}, which considers the global gain 
of the path grades, instead of the individual/local ones). 
Thus this simulation highlights the ability of the cooperative controller  
to determine locally sub-optimal tasks for SaR robots, in order to maximise the global mission performance. 
\Cref{fig:3scan} shows that by spreading out the SaR robots over the environment, 
the cooperative controller achieves an overall scan certainty that is $18.7\%$ 
larger than that of the selfish controller.%

\noindent
\textbf{Case 4: Exploitation of sensor accuracies:} 
Consider $2$ SaR robots in an environment of size $15\times 15$, 
where the robots should scan a partially known environment with two sub-areas 
$E'_1, E'_2\subset E$ (see \Cref{fig:case4}). 
The scan certainty at the initial time step for sub-areas $E'_1$ and $E'_2$ 
is $0$ and $0.3$, respectively, and for all cells of $E$ outside these two sub-areas is $0.9$. 
Figures~\ref{fig:4track} and \ref{fig:4scan} show, respectively, 
the paths taken by the robots for $10$ time steps using both cooperative and selfish 
controllers and the change in the total scan certainty in time.%

Based on \Cref{fig:4track}, with the selfish controller both SaR robots move to sub-area $E'_1$ 
to yield the highest gain in the scan certainty. 
With the cooperative controller, however, the robots move to sub-areas $E'_1$ and $E'_2$. 
Since SaR robot $\textrm{a}_1$ is closer to sub-area $E'_1$ and has a higher sensor accuracy, 
it is sent to sub-area $E'_1$ by the cooperative controller to yield a larger 
overall scan certainty. 
Based on \Cref{fig:4scan} the overall scan certainty of 
the cooperative controller is $29.8\%$ larger than that of the selfish controller.%

\noindent
\textbf{Case 5. Combined scenario:} 
Consider $2$ SaR robots, $6$ victims, and a set of obstacles in an environment of size $30\times 15$, 
where the robots should scan a partially known environment with $7$  sub-areas $E'_1,\ldots, E'_7\subset E$ (see \Cref{fig:combi_layout}). 
The scan certainties at the initial time step for sub-area $E'_1$ is $0.7$, for sub-area $E'_2$ is $0.4$, 
for sub-areas $E'_3, E'_4, E'_5, E'_6$ is $0.2$, for 
sub-area $E'_7$ is $0.1$, and for all cells of $E$ outside these seven sub-areas is $0.5$. 
The selfish and cooperative paths of both SaR robots 
are shown in \Cref{fig:combitrack} for $35$ time steps. Additionally, \Cref{tab:combi_vic} shows the health state of the victims at the end of the simulation, the number of times a victim has been visited, and the  time step when each victim was first detected. The change in total scan certainty in time  
is  also illustrated in \Cref{fig:combiscan}.%

\Cref{fig:combitrack} shows that with the selfish controller, 
both SaR robots visit victim $\mathrm{v}_3$, and then move towards the southwest 
quadrant of the SaR environment, 
where they individually visit victim $\mathrm{v}_2$. 
While $2$ victims are visited doubly by the robots, $4$ victims remain undetected and $1$ victim 
deceases. 
With the cooperative controller, SaR robots $\textrm{a}_1$ and $\textrm{a}_2$ visit 
victims $\mathrm{v}_4$ and $\mathrm{v}_3$, respectively. Next they explore different sub-areas of the environment 
and detect additional victims $\mathrm{v}_2$, $\mathrm{v}_5$, and $\mathrm{v}_6$. 
At the end only $1$ victim remains undetected and no victim is deceased.%

The SaR system detects more victims with the cooperative controller, 
and no victim is visited twice, implying the victim search efficiency. 
Moreover, the overall scan certainty for the cooperative controller (see \Cref{fig:combiscan})  
is $27.6\%$ larger than that for the selfish controller.%
%

%



\section{Conclusions and Topics for Future Research}
\label{sec:concl}

Autonomous multi-robot systems are expected to map unknown search-and-rescue (SaR) 
environments in a fast and effective way. 
We have introduced a novel approach for coordinated mission planning of multi-robot 
systems for multi-objective (combined coverage and target-oriented) SaR. 
The developed control approach effectively incorporates non-homogeneous imperfect 
perception capabilities of the sensors of different robots in order to improve 
their performance with respect to the victim detection and area coverage.%

The key contributions of the paper are two-fold: in multi-agent control systems 
and in search-and-rescue (SaR) robotics. 
From the point-of-view of multi-agent control systems, we propose 
a novel control architecture and formulation that exploit the imperfect 
perception capabilities of agents, coordinate their decisions, and 
provide a balanced trade-off among various control objectives 
in a computationally efficient way. 
As is also supported by our simulation results, the developed control system benefits 
from both computational efficiency of decentralised control methods and 
global vision of centralised control approaches. 
Additionally, the supervisory level improves the global control performance 
based on a predictive and optimal computation scheme, while local controllers 
independently steer the agents. Therefore, although the performance will expectedly 
degrade, the function of the multi-agent control system is robust to the failure of 
this centralised controller. 
Furthermore, the integrated formulation proposed in this research allows to 
incorporate both expert knowledge (via the fuzzy logic control systems)  
and the optimality and predictive capabilities of model predictive control (MPC) 
into the decision making of autonomous robots. 
These contributions are significant for SaR applications, because existing control methods 
are mainly focused on either coverage or target-oriented SaR. Moreover, MPC, 
which is a precise control method that systematically handles state and input constraints 
and that can provide robustness to SaR uncertainties, has been ignored in the literature for 
the crucial task of area coverage in SaR. 
Our novel approach and formulation for multi-agent control 
systems enables MPC to provide all its strong points for, not only target-oriented, but 
also coverage-oriented SaR.%

We have compared the performance of the resulting cooperative control system with those of 
a decentralised selfish control system that excludes the MPC controller, a pure MPC controller, 
an ant-colony-based controller, and an exhaustive random search controller. 
In $20$ simulated scenarios with randomly positioned obstacles and victims, 
the hierarchical control approach showed the best performance in terms of 
victim detection efficiency and area coverage. 
Moreover, $5$ structured scenarios were designed to simulate conflicting scenarios 
and to illustrate the importance of the proposed mathematical formulations in application. 
The results proved 
that in case of conflicts, the proposed hierarchical controller significantly outperforms the decentralised controller 
with a comparable computation time. 
Moreover, the hierarchical controller successfully exploits the non-homogeneous 
perception capabilities of robots, which improves the overall performance.

In the future, more detailed models that consider the behaviour, physical capabilities, and intentions of 
victims for their movement patterns can be considered. 
Furthermore, a systematic discussion and evaluation of the robustness of the 
proposed control approaches with respect to various sources of uncertainties, 
especially uncertainties in the movement of victims, is a topic of interest for future research. 
Moreover, in addition to non-homogeneous perception capabilities, 
differences in the speed, degrees of freedom, computational capacity, tasks, and maneuverability 
of search-and-rescue robots can be considered. 
Additionally, combining autonomous learning methods within the proposed architecture is an interesting topic for future research. 
While using a learning-based approach alone may correspond to some risks for search-and-rescue applications, 
including such algorithms in a combined framework, similar to the one proposed in this paper, 
can result in a promising performance with adaptability capabilities. 
Finally, in real-life implementations the large size of SaR environments increases the computational burden of the supervisory MPC controller. To address this issue and also to mitigate the risk of performance degradation due to failure of the supervisory control level, a similar control architecture with more levels of control may be proposed. Thus, between the supervisory control level and the steering local control level, extra levels of control with several distributed MPC controllers are considered, where each MPC controller supervises a combination of local sub-areas.%

\section{Statements and Declarations}
\subsubsection*{Funding}

This research has been supported by the NWO Talent Programme Veni project ``Autonomous drones flocking for search-and-rescue'' (18120), which has been financed by the  Netherlands Organisation for Scientific Research (NWO).

\subsubsection*{Competing Interests}
The authors declare that they have no financial or non-financial conflict of interest. 

\subsubsection*{Data Availability}

The data points, files, and codes for creating the figures represented in the results of this article are available online at https://figshare.com/s/9762330a7473363433ab.

\subsubsection*{Author Contributions}

Author C.\ de Koning   
contributed to designing and implementing the experiments. 
Authors C.\ de Koning and A.\ Jamshidnejad 
contributed to the analysis and interpretation of the results, 
development of the theoretical contributions, 
and composition of the manuscript. 
Author C.\ de Koning prepared the first draft of the manuscript. 
Author A.\ Jamshidnejad supervised the study
design, has critically reviewed and edited the manuscript and 
has prepared the final version of the paper. 
Both authors have approved the final version of the manuscript.

\subsubsection*{Ethics approval}
Not applicable.

\subsubsection*{Consent to participate}
Not applicable.

\subsubsection*{Consent for publication}
Not applicable.


\bibliography{sn-bibliography}



\end{document}